\pdfoutput=1

\documentclass[11pt]{article}

\usepackage[final]{latex/acl}

\usepackage{times}
\usepackage{latexsym}

\usepackage{xcolor}

\newcommand{\xref}[1]{\S\ref{#1}}

\usepackage{booktabs}%

\usepackage[T1]{fontenc}

\usepackage[utf8]{inputenc}

\usepackage{microtype}

\usepackage{inconsolata}

\usepackage{graphicx}

\usepackage{subcaption}
\usepackage{mwe}

\usepackage{multicol}
\usepackage{multirow}
\usepackage{tabularx}
\usepackage{enumitem}
\usepackage{amssymb}
\usepackage{pifont}
\newcommand{\cmark}{\ding{51}}%
\newcommand{\xmark}{\ding{55}}%

\usepackage{makecell}
\usepackage{adjustbox}

 \title{Reading Between the Lines: The One-Sided Conversation Problem}

\author{
 \textbf{Victoria Ebert,\textsuperscript{1}}
 \textbf{Rishabh Singh,\textsuperscript{1}}
 \textbf{Tuochao Chen,\textsuperscript{1,3}}
\\
 \textbf{Noah A. Smith,\textsuperscript{1, 2}}
 \textbf{Shyamnath Gollakota\textsuperscript{1,3}}
\\
\\
 \textsuperscript{1}Paul G. Allen School of Computer Science \& Engineering, University of Washington,
 \\
 \textsuperscript{2}Allen Institute for Artificial Intelligence
\\
\textsuperscript{3}Hearvana AI\\
\texttt{\{ebertv,rissingh,tuochao,nasmith,gshyam\}@cs.washington.edu }
}

\newif\ifshowcomments

\newcommand{\squishlist}{\begin{itemize}[itemsep=1pt,parsep=2pt,topsep=3pt,partopsep=0pt,leftmargin=0em, itemindent=1em,labelwidth=1em,labelsep=0.5em]}

\newcommand{\squishend}{\end{itemize}}

\showcommentsfalse

\begin{document}
\maketitle

\begin{abstract}
    
Conversational AI is  constrained in many real-world settings where only one side of a dialogue can be recorded.
We formalize the {\it one-sided conversation problem (1SC)}: inferring and learning from only one side of a conversation.  We study two tasks: (1) reconstructing the missing speaker's turns
and (2) generating summaries from one-sided transcripts. Evaluating
models on MultiWOZ, DailyDialog, SpokenWOZ and Candor with both human A/B testing and LLM-as-a-judge metrics, we find that additional context improves reconstruction,
and while large models generate promising reconstructions with prompting, smaller models require finetuning. Further, high-quality summaries can be generated without reconstructing missing turns. We present 1SC as a novel challenge and report promising results that mark a step toward privacy-aware conversational AI.


\end{abstract}

\section{Introduction}

\begin{figure}[t]
  \includegraphics[width=\columnwidth]{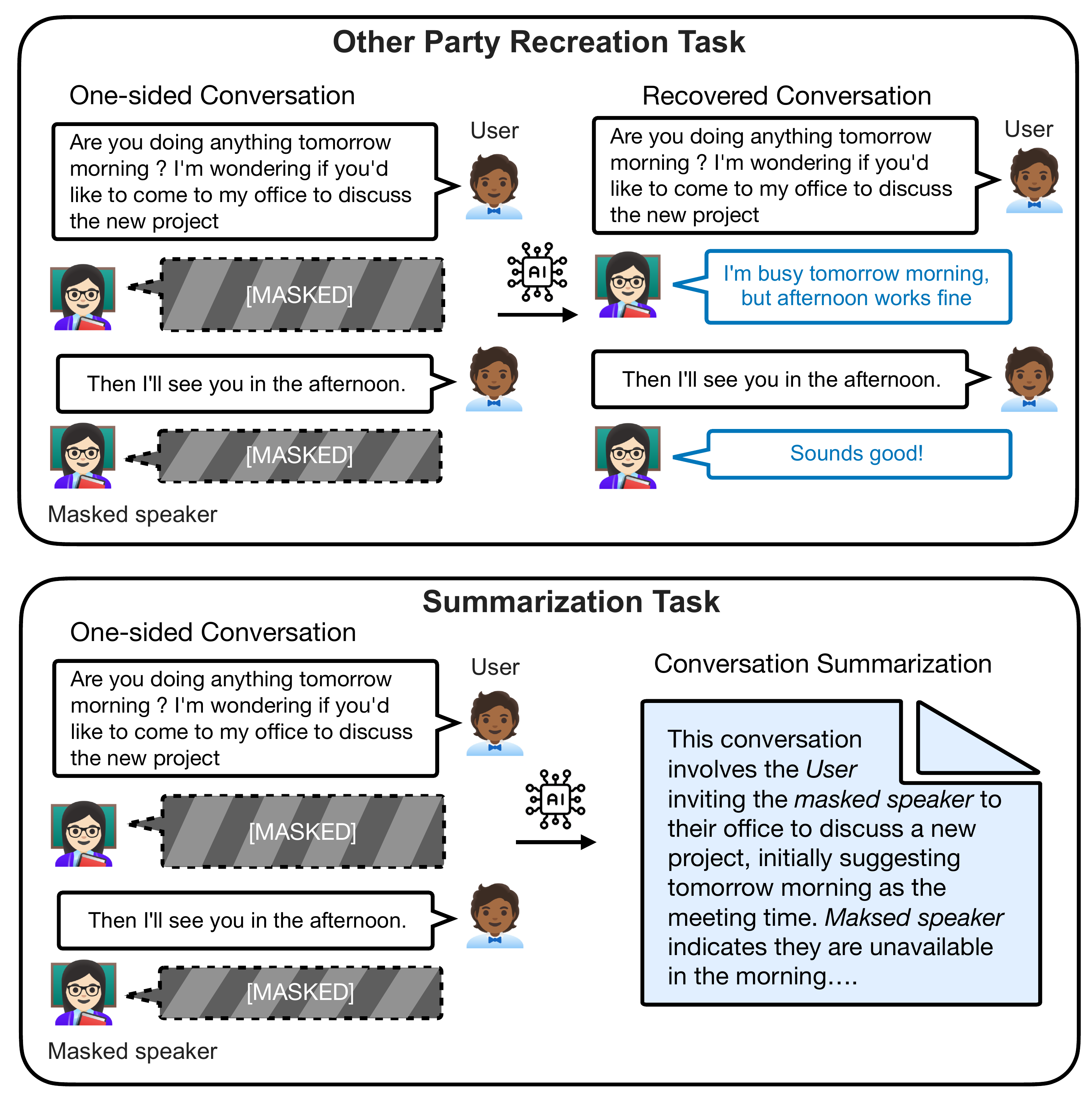}
  \vskip -0.1in
  \caption{We introduce the one-sided conversation (1SC) problem: making inferences from only one side of a conversation transcript. We focus on reconstruction of the missing content and creating summaries of the whole  one-sided  conversation.}
  \label{fig:figure1}
  \vskip -0.15in
\end{figure}

Two converging trends are setting the stage for always-available, context-aware assistants: the rise of conversational AI and the spread of augmented reality devices such as smart glasses and in-ear augmentation~\cite{bubble,10.1145/3613904.3642057}. Conversational agents now power virtual assistants, call centers, and telemedicine platforms~\cite{10.1145/3654777.3676430,medicalnature}, while new human-augmentation applications are emerging, including real-time meeting summarization~\cite{summary-1}, personalized social coaching~\cite{10.1145/3706598.3713606}, and discreet in-ear guidance~\cite{llamapie}. In these settings, conversational AI extends human cognition, capturing and augmenting dialogue to enhance memory, efficiency, and accessibility~\cite{10.1145/3613904.3642450}.

A fundamental barrier, however, remains: in many real-world scenarios, only {one side of a conversation} is available for processing. This asymmetry stems from both technical and legal constraints. For instance, call centers and telemedicine platforms often record only the agent’s or patient’s transcripts for compliance. Similarly, smart glasses and in-ear assistants have the ability to capture only the wearer’s speech~\cite{clearbuds,proactiveagent-emnlp} to preserve privacy and satisfy jurisdictional requirements. In the United States, recording laws vary: one-party states allow recording with a single participant’s consent, while all-party states (California, Delaware, Florida, Illinois, Maryland, Massachusetts, Montana, Nevada, New Hampshire, Pennsylvania, Washington at time of publication) require consent from every speaker~\cite{matthiesen2019laws}.\footnote{U.S. federal law (18 U.S.C. §2511) prohibits the intentional interception of oral, wire, or electronic communications, except as authorized, for example via consent. 
~\cite{recordinglaw}.
In the EU, 
organizations using tools to record or transcribe conversations must establish a lawful basis (often explicit consent)~\cite{Nautsch_2019,eu-1,eu-2}.}  Thus, dialogue systems built for analysis or augmentation often have access to only one-sided input.

We define the {\it one-sided conversation problem}  (1SC): inferring and learning from a dialogue when only one speaker’s utterances are observed. Unlike existing text-infilling tasks that fill short gaps, 1SC requires reasoning over an absent speaker’s entire dialogue. Solutions must minimize hallucinations, avoiding fabricated facts such as names or dates, and instead provide principled ways to make one-sided data useful, especially for the participant whose speech is recorded.  To our knowledge, 1SC has not yet been studied in the literature. The most directly related work to the one-sided conversation problem is discussed throughout the paper; see \S\ref{sec:appen:related} for additional connections.\footnote{{The one-sided conversation problem is fundamentally a data privacy problem. Rather than advancing role-playing dialogue agents, our paper explores a realistic, privacy-aware setting for developing AI systems that support a human in real dialogue, without exposing the other parties to the AI.}}

We investigate two tasks with 1SC: reconstructing the missing speaker's turns in an online manner, given varying amounts of context (\S\ref{sec:opr}), and generating summaries from one-sided transcripts, with or without reconstructed turns (\S\ref{sec:summ}). The first supports real-time applications such as proactive guidance~\cite{llamapie} and conversation-aware intelligence~\cite{10.1145/3613904.3642450}, where turn-by-turn inference is essential. The second enables post-hoc analysis, producing faithful summaries of conversations even when one side is missing.

In \S\ref{sec:real}, we evaluate prompting and finetuned models on two text based dialogue datasets: MultiWOZ (task-oriented) and DailyDialog (open-domain) and two audio-based datasets: SpokenWOZ (task-oriented) and Candor (open-domain). To assess  quality, we introduce an evaluation framework using \textsc{GPT-4o} as a judge, scoring semantic preservation, intent alignment, and hallucination of specific details. Further, an A/B test with 16 human participants assesses whether humans can distinguish ground-truth dialogues from LM-generated one-sided reconstructions.

 Our findings reveal several insights:
\begin{enumerate}
[itemsep=1pt,parsep=2pt,topsep=3pt,partopsep=0pt,leftmargin=0em, itemindent=2em,labelwidth=1em,labelsep=0.5em]
\item Compared to using only preceding context, access to the turn immediately following a masked turn improves reconstruction. 
\item Information about the length of a masked turn improves reconstruction.
\item Large pretrained models
can generate plausible reconstructions out-of-the-box; even with finetuning, smaller models 
do not achieve the same capabilities. 
\item  High-quality summaries can be produced directly from one-sided input \emph{without} reconstructing missing turns. 
\item While models struggle with reconstruction of speech-based transcriptions, the summaries produced from these conversations are on par with those of text-based conversations. 
\end{enumerate}

Our results expose limitations of current models and suggest future directions in dialogue infilling, controllable generation, and privacy-aware AI.

By formalizing the 1SC problem, studying two of its tasks, and establishing evaluation metrics, our work lays the foundation for systems that can operate robustly under the incomplete yet realistic conditions of human conversations.

We release our code and data at \url{https://github.com/ebertv/onesided}.






\section{Other Party Recreation in 1SC} \label{sec:opr}

We first consider explicit recreation of the missing party's dialog turns using language models, focusing initially on text-based datasets.

\subsection{Motivation}
The key motivation for explicit online utterance reconstruction is the emergence of a new class of proactive conversational assistants designed to support people during live interactions. Unlike summarization, which can be performed retrospectively after a conversation concludes, proactive guidance requires the assistant to track the dialogue as it unfolds and intervene with timely, contextually appropriate support. Recent work on augmented-reality–based conversational guidance systems for smart glasses, earbuds and hearing aids, such as Memoro~\cite{10.1145/3613904.3642450}, LLAMAPIE~\cite{llamapie}, and Overhearing LLM Agents~\cite{zhu2025overhearingllmagentssurvey}, exemplifies this trend.

Inexact utterance reconstruction can introduce noise into a system, and therefore is not useful in all cases. However, in some applications, exact reconstruction may not be essential. For example, in role-playing scenarios, reconstruction can be used to generate plausible conversational partners.  Such scenarios arise in training, e.g., doctor-patient interactions for residents or reconstructed conversations from experienced call center agents for trainees.  \emph{Multiple} plausible reconstructions are useful in forensic settings, similar to modern lip-reading techniques that generate several candidate transcripts. These kinds of applications capture the value of reconstruction even when accuracy is imperfect. Further motivation can be found in \S\ref{sec:appen:motiv}.

\subsection{Task Formulations} \label{sec:opr:task}
\begin{figure}[t!]
        \centering
        \begin{subfigure}[b]{0.46\columnwidth}
            \centering
            \includegraphics[width=\columnwidth]{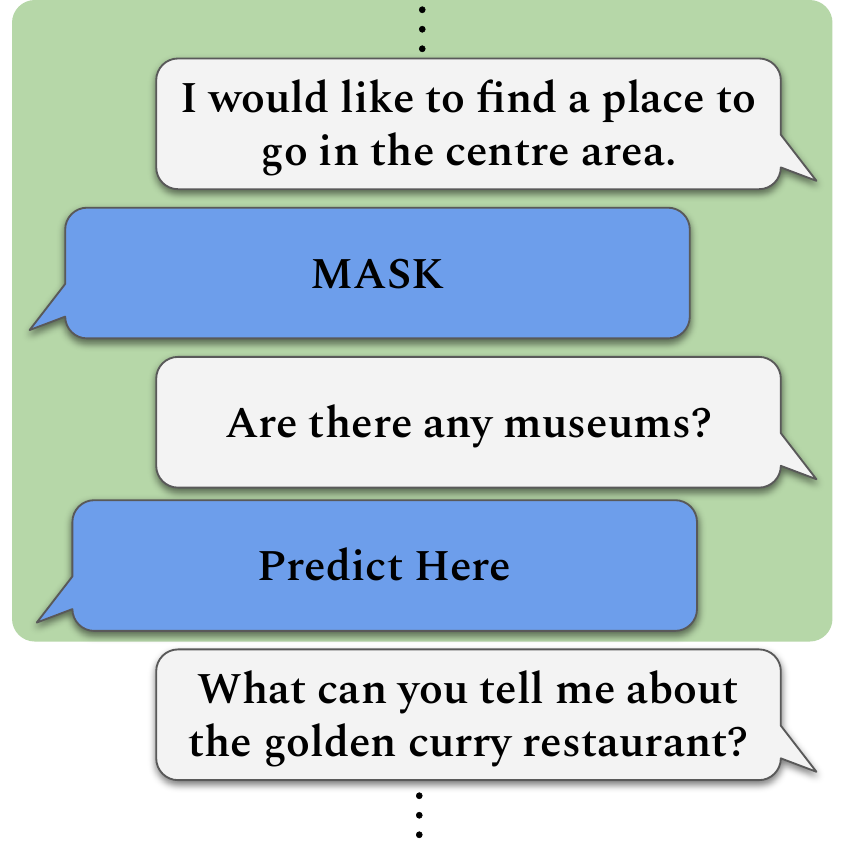}
            \caption{{\small Baseline}}    
            \label{fig:exbase}
        \end{subfigure}
        \hfill
        \begin{subfigure}[b]{0.46\columnwidth}  
            \centering 
            \includegraphics[width=\columnwidth]{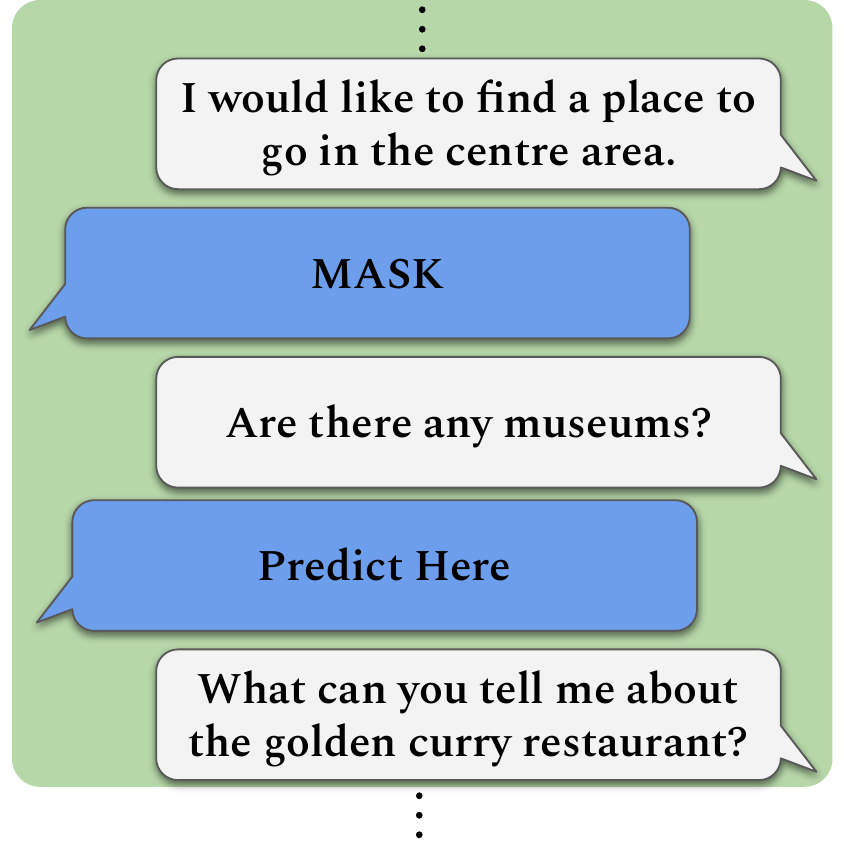}
            \caption{{\small Include Turn N+1}}    
            \label{fig:exft}
        \end{subfigure}
        \begin{subfigure}[b]{0.46\columnwidth}   
            \centering 
            \includegraphics[width=\columnwidth]{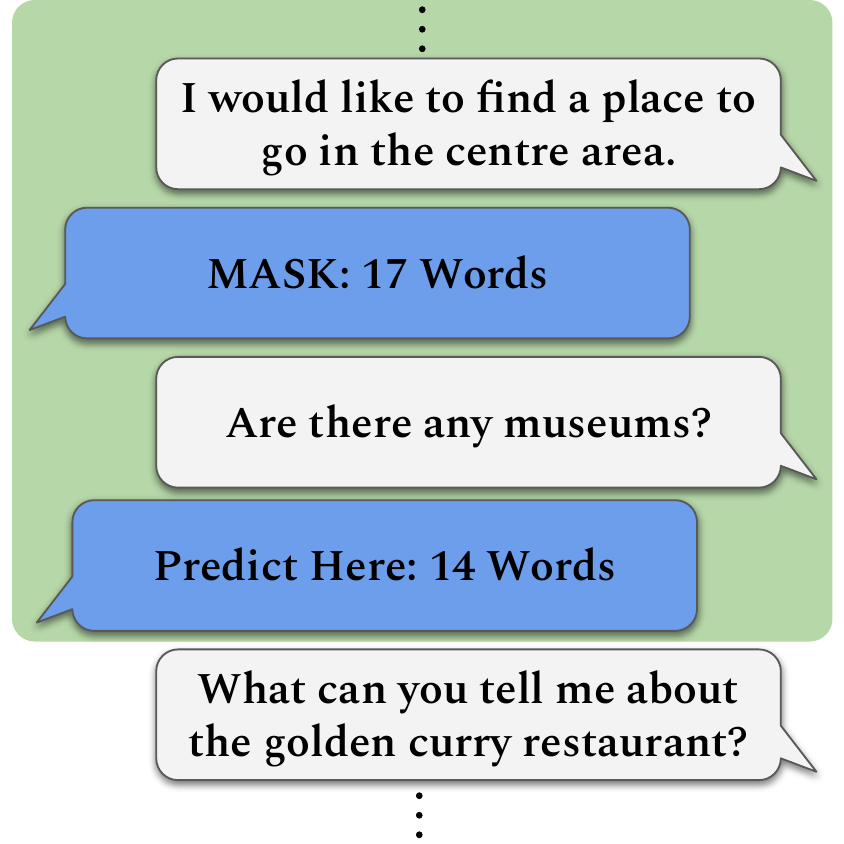}
            \caption{{\small Include Turn Lengths}}    
            \label{fig:extl}
        \end{subfigure}
        \hfill
        \begin{subfigure}[b]{0.46\columnwidth}   
            \centering 
            \includegraphics[width=\columnwidth]{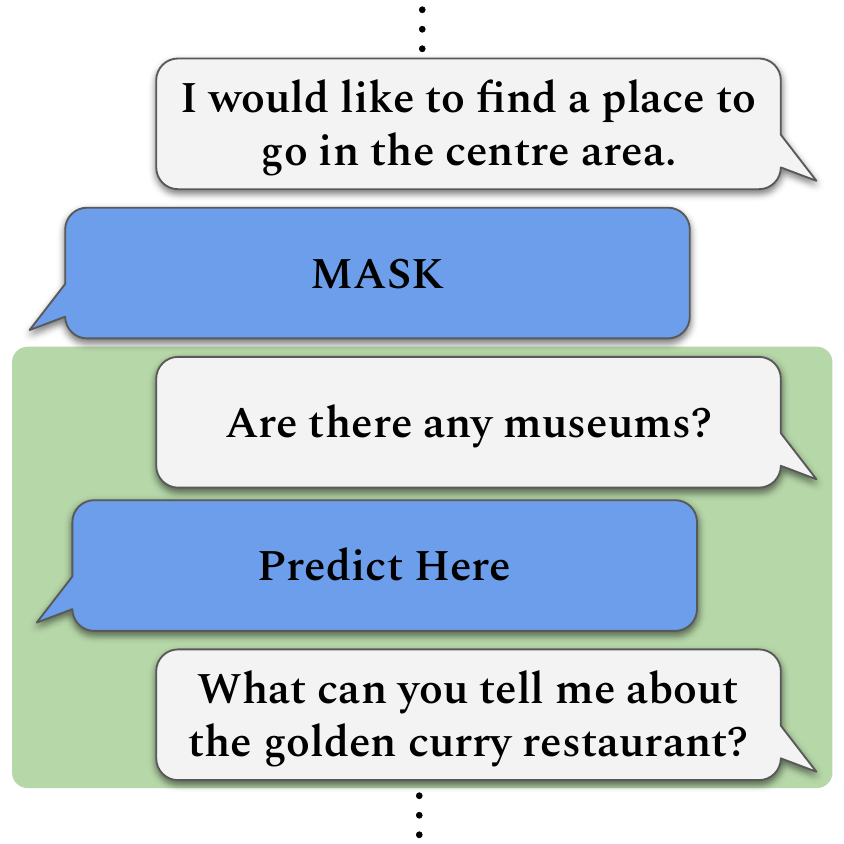}
            \caption{{\small Limited Context}}    
            \label{fig:3t}
        \end{subfigure}
        \vskip -0.05in       
        \caption{Different levels of context considered for other party reconstruction. Context is marked by green boxes; this example predicts turn 4. (a) gives the whole conversation up to turn 4. (b) includes turn 5. (c) includes the length of each utterance. (d) gives only local context (turns 3 and 5).}
        \vskip -0.15in
        \label{fig:examples}
    \end{figure}

In our two-party setting, the goal is to infer  one person's side of a conversation given only the other person's utterances. We refer to the observed side as the user and the side to be inferred as the masked speaker. This task targets  applications like  proactive guidance~\cite{llamapie} and conversation-aware intelligence~\cite{10.1145/3613904.3642450}, where real-time turn-by-turn inference is essential. We focus on single-turn prediction, as  experiments showed that prompting a LM to fill multiple turns was ineffective (see Table~\ref{tab:fullconvo}). Multiple levels of context around each turn are considered (Fig.~\ref{fig:examples}).

{\bf Full one-sided prior context.}  We explore a setting where each turn from the masked speaker is predicted immediately after the preceding user utterance (Fig.~\ref{fig:exbase}), referring to the predicted turn as {\it Turn N}. In this setup, the conversation with one speaker masked is provided up to Turn N. We examine two variations:

\squishlist
    \item \textit{Include Turn N+1}. 
    Incorporates the user’s next utterance to assess whether additional context improves Turn N prediction while preserving slightly-delayed but still online inference (Fig.~\ref{fig:exft}).
    
    \item \textit{Include Turn Lengths}. Tests whether information about the duration of time between user turns can help predict the masked speaker’s response. In text-based datasets, we use word count as an indirect proxy for timing  (Fig.~\ref{fig:extl}).

    \squishend

{\bf Local one-sided context.} We analyze a more context-limited setting where the system is given only Turn N-1, Turn N, and Turn N+1 (a total of three turns; two from the user and the middle from the masked speaker; Fig.~\ref{fig:3t}). This restricts the system to using only the immediate surrounding information as context but is computationally cheaper, especially in longer conversations.

\subsection{Datasets} \label{sec:opr:data}
We use  MultiWOZ \cite{budzianowski2020multiwozlargescalemultidomain}, SpokenWOZ \cite{10.5555/3666122.3667821}, DailyDialog \cite{li-etal-2017-dailydialog}, and the Candor Corpus \cite{doi:10.1126/sciadv.adf3197}, for evaluation. MultiWOZ and DailyDialog are human-written; MultiWOZ contains task-oriented dialogues and DailyDialog focuses on everyday conversations (often for English learning). SpokenWOZ is an audio dataset derived from speakers playing the roles of users and agents, where the agent has access to the same database used to construct MultiWOZ. Transcripts of these conversations create a task-oriented dataset that contains many of the the nuances of spoken dialogues. Candor is also a spoken dataset, derived from video-chat transcriptions of unscripted conversations. Candor features interruptions and single-word responses; it is the most challenging dataset in our study. We focus initially on the text-based conversations from DailyDialog and MultiWOZ (\S\ref{sec:opr}--\ref{sec:summ}), and then move to the realistic setting with SpokenWOZ and Candor (\S\ref{sec:real}).

We also use SODA \cite{kim2023sodamillionscaledialoguedistillation} for a synthetic and large-scale dataset of social dialogues. However, since our focus is human dialogue, we use only its training and validation splits for finetuning a predictive model, and we perform no evaluation using SODA.

Further details of all datasets, including data size and partitioning, are given in Table~\ref{tab:splits} in \S\ref{sec:appen:data}. 

\subsection{Methods for Other Party Recreation}
\subsubsection{Finetuning}

We finetune Llama-3.2-1B \cite{grattafiori2024llama3herdmodels} 
on the limited context task (Fig.~\ref{fig:3t}). Finetuning is performed auto-regressively, following prior work on finetuning language models for the related task of infilling~\cite{donahue-etal-2020-enabling}. 

\subsubsection{Prompting}

We prompt Claude-4-Sonnet and Llama-3.2-1B-Instruct. We use Llama's  instruction-tuned version for prompting, as it better handles task-oriented responses~\cite{grattafiori2024llama3herdmodels}. 
Llama is prompted only in the three-turn setup to match the finetuned setting, whereas Claude is prompted for both complete-context and three-turn settings. All models receive the same prompt and are instructed to predict a natural response that preserves context details, using ``xxxx'' as a placeholder for unknown information. They maintain the conversation’s tone and information density, with future turns intended solely for context, not as knowledge available to the masked speaker. 

Few-shot examples are included, and instructions are reiterated after the masked conversation (see~\xref{apdx:prompt:infill}). The models  receive system-level instructions, instructing them to stay focused and respond only with the system output. 

\subsection{Experiments}\label{sec:opr:eval}

\begin{table}[]
    \centering
    \resizebox{\columnwidth}{!}{
    \begin{tabular}{cccccc}
         & & \makecell{Ground\\Truth} & Model & Tie & \makecell{No\\Majority}\\
        \midrule[2pt]
        \multirow{2}{*}{\makecell{Daily\\Dialog}} & Claude & 28\% & 32\% & 20\% & 20\%\\
        & Llama & 68\% & 8\% & 20\% & 4\%\\
        \midrule
        \multirow{2}{*}{\makecell{Multi\\WOZ}} & Claude & 12\% & 56\% & 20\% & 12\%\\
        & Llama & 40\% & 12\% & 44\% & 48\%\\
        \midrule
        \midrule
        SpokenWOZ & Claude & 16.0\% & 29.3\% & 32.0\% & 22.7\%\\
        \midrule
        Candor & Claude & 11.0\% & 40.2\% & 25.6\% & 23.2\%\\
        \bottomrule
    \end{tabular}}
    \vskip -0.05in
    \caption{Percentages of conversations where a clear majority of judges selected each response option. In some cases there was no clear majority, e.g., if six judges were split equally among ``Ground Truth,'' ``Model,'' and ``Tie''; we count those cases in the right-most column.}
    \vskip -0.15in
    \label{tab:humanrecon}
\end{table}

\subsubsection{A/B testing with Human Evaluators}\label{sec:opr:eval:hum}

To evaluate reconstructed turns, we conducted an A/B test with 16 human judges. Judges were shown three-turn dialogue contexts and presented with two candidate responses for the masked turn generated from our local-context setting. They were asked to select the response that better fit the context; annotators were allowed to select ``neither,'' but were instructed that this choice should be used sparingly. (see \xref{apdx:annotators} for further details).

When given the conversational dialogues from DailyDialog $(n=50)$, annotators were more likely to prefer reconstructions from Claude $(n=25)$ to those from the ground truth dataset, but the ground truth was heavily preferred over reconstructions from the finetuned Llama model $(n=25)$. On task-oriented dialogues from MultiWOZ $(n=50)$, the preference for Claude reconstructions $(n=25)$ over the ground truth was stronger and the distaste for finetuned Llama reconstructions $(n=25)$ over the ground truth was smaller. The top two rows of Table~\ref{tab:humanrecon} give a full breakdown of human judgments. 
Overall results indicate no clear preference for ground truth over reconstructions, particularly with larger models. 

\subsubsection{Automated Evaluation} \label{sec:opr:eval:llm}
Given the large scale of our datasets, and multiple variants to be evaluated,  we  use LLM-as-a-judge methods for more fine-grained evaluation \cite{zheng2023judgingllmasajudgemtbenchchatbot, kim2024prometheusinducingfinegrainedevaluation}.

\vskip 0.05in\noindent{\bf Metrics.} We evaluate using both rubric scores as well as precision-recall metrics.

{\it Rubric scores.}  We ask the evaluator model to reason about two dialogues where one is the actual dialogue at a specific turn, and the other was given by the predictor model. Following best practices, we use a separate model for evaluation (\textsc{GPT-4o}) which provides detailed reasoning and an ordinal score (values 1--5) for each of five criteria:

\definecolor{cA}{HTML}{EDF4FD}
\definecolor{cB}{HTML}{D6E8FA}
\definecolor{cC}{HTML}{BFDCF7}
\definecolor{cD}{HTML}{A8D0F4}
\definecolor{cE}{HTML}{91C4F1}
\definecolor{cF}{HTML}{7AB8EE}
\definecolor{lA}{HTML}{FDF1DC}
\definecolor{lB}{HTML}{FAD99A}

\begin{table*}[ht]
    \centering
    \renewcommand{\arraystretch}{0.9}
    \setlength{\tabcolsep}{3.5pt}
    \resizebox{\textwidth}{!}{
    \footnotesize
    \begin{tabular}{lccccccccccc}
    \toprule[2pt]
    Dataset & N+1 & \makecell{Turn\\Len.} & \makecell{``xxxx''\\Instr.} & \makecell{Full Prior\\Context} 
    & \makecell{\textbf{P}rompted/\\\textbf{F}inetuned} & \makecell{\textbf{C}laude/\\ \textbf{L}lama} 
    & \makecell{Seman.\\Sim. ($\uparrow$)} & \makecell{Intent\\Pres. ($\uparrow$)} 
    & \makecell{Context.\\Approp. ($\uparrow$)} & \makecell{Summ.\\Align. ($\uparrow$)} 
    & \makecell{Anti-\\Halluc. ($\uparrow$)} \\
    \midrule[2pt]
    
    \multirow{8}{*}{\makecell{Daily\\Dialog}} 
    & \xmark & \xmark & \xmark & \cmark & \textsc{P} & \textsc{C} & 1.79 (0.95) & 2.65 (1.34) & 3.11 (1.09) & 1.80 (0.96) & 4.05 (1.61)\\
    & \xmark & \xmark & \cmark & \cmark & \textsc{P} & \textsc{C} & 1.85 (0.92) & 2.75 (1.37) & 3.32 (1.08) & 1.89 (0.94) & 4.60 (0.99)\\
    & \cmark & \xmark & \cmark & \cmark & \textsc{P} & \textsc{C} & 2.34 (1.07) & 3.42 (1.32) & 3.60 (1.06) & 2.39 (1.13) & 4.65 (0.92)\\
    & \xmark & \cmark & \cmark & \cmark & \textsc{P} & \textsc{C} & 2.14 (1.18) & 2.97 (1.49) & 3.77 (1.10) & \textbf{2.81 (1.29)} & 4.66 (0.90)\\
    & \cmark & \cmark & \cmark & \cmark & \textsc{P} & \textsc{C} & \textbf{2.63 (1.28)} & \textbf{3.60 (1.36)} & \textbf{3.80 (1.05)} & 2.71 (1.31) & \textbf{4.77 (0.76)}\\
    & \cmark & \xmark & \cmark & \xmark & \textsc{P} & \textsc{C} & 2.12 (1.11) & 2.95 (1.46) & 3.44 (1.12) & 2.17 (1.15) & 4.60 (0.96)\\
    & \cmark & \xmark & \cmark & \xmark & \textsc{P} & \textsc{L} & 1.21 (0.47) & 1.41 (0.76) & 1.57 (0.85) & 1.20 (0.45) & 1.87 (1.16)\\
    & \cmark & \xmark & \cmark & \xmark & \textsc{F} & \textsc{L} & 1.38 (0.88) & 1.80 (1.27) & 1.64 (1.03) & 1.39 (0.90) & 2.49 (1.60)\\
    
    \midrule
    
    \multirow{8}{*}{\makecell{Multi\\WOZ}}  
    & \xmark & \xmark & \xmark & \cmark & \textsc{P} & \textsc{C} & 2.50 (1.09) & 3.54 (1.38) & 3.39 (1.19) & 2.44 (1.09) & 3.14 (1.91)\\
    & \xmark & \xmark & \cmark & \cmark & \textsc{P} & \textsc{C} & 2.50 (1.01) & 3.60 (1.33) & 3.69 (1.08) & 2.50 (1.04) & 4.71 (0.77)\\
    & \cmark & \xmark & \cmark & \cmark & \textsc{P} & \textsc{C} & 2.64 (1.00) & 3.87 (1.21) & 3.76 (1.01) & 2.66 (1.03) & 4.59 (0.92)\\
    & \xmark & \cmark & \cmark & \cmark & \textsc{P} & \textsc{C} & 2.84 (1.27) & 3.81 (1.36) & 3.95 (1.12) & 2.81 (1.29) & 3.98 (1.65)\\
    & \cmark & \cmark & \cmark & \cmark & \textsc{P} & \textsc{C} & \textbf{2.96 (1.20)} & \textbf{4.06 (1.17)} & \textbf{4.07 (0.98)} & \textbf{3.00 (1.24)} & \textbf{4.74 (0.75)}\\
    & \cmark & \xmark & \cmark & \xmark & \textsc{P} & \textsc{C} & 2.59 (1.06) & 3.70 (1.29) & 3.76 (1.05) & 2.61 (1.09) & 4.67 (0.82)\\
    & \cmark & \xmark & \cmark & \xmark & \textsc{P} & \textsc{L} & 1.63 (0.66) & 1.98 (0.95) & 2.06 (0.90) & 1.60 (0.61) & 2.24 (1.13)\\
    & \cmark & \xmark & \cmark & \xmark & \textsc{F} & \textsc{L} & 1.98 (1.12) & 3.13 (1.42) & 2.39 (1.17) & 1.98 (1.11) & 2.46 (1.46)\\

    \midrule
    \midrule

    \multirow{5}{*}{\makecell{Spoken\\WOZ}}
    & \xmark & \xmark & \xmark & \cmark & \textsc{P} & \textsc{C} & 1.62 (0.99) & 2.19 (1.39) & 2.13 (1.24) & 1.60 (0.98) & 1.91 (1.18)\\
    & \xmark & \xmark & \cmark & \cmark & \textsc{P} & \textsc{C} & 1.67 (0.99) & 2.22 (1.38) & 2.22 (1.24) & 1.66 (1.01) & 2.18 (1.32)\\
    & \cmark & \xmark & \cmark & \cmark & \textsc{P} & \textsc{C} & 1.85 (1.11) & 2.51 (1.47) & 2.30 (1.27) & 1.85 (1.11) & 2.28 (1.33)\\
    & \xmark & \cmark & \cmark & \cmark & \textsc{P} & \textsc{C} & 2.07 (1.32) & 2.65 (1.61) & 2.67 (1.42) & 2.11 (1.37) & 3.14 (1.54)\\
    & \cmark & \cmark & \cmark & \cmark & \textsc{P} & \textsc{C} & \textbf{2.29 (1.36)} & \textbf{2.96 (1.60)} & \textbf{2.76 (1.36)} & \textbf{2.31 (1.40)} & \textbf{3.22 (1.52)}\\
    
    \midrule
    
    \multirow{5}{*}{Candor}
    & \xmark & \xmark & \xmark & \cmark & \textsc{P} & \textsc{C} & 1.18 (0.48) & 1.48 (0.82) & 1.86 (0.92) & 1.18 (0.46) & 1.99 (1.2)\\
    & \xmark & \xmark & \cmark & \cmark & \textsc{P} & \textsc{C} & 1.19 (0.50) & 1.53 (0.85) & 1.94 (0.95) & 1.19 (0.49) & 2.17 (1.31)\\
    & \cmark & \xmark & \cmark & \cmark & \textsc{P} & \textsc{C} & 1.24 (0.58) & 1.58 (0.97) & 1.82 (0.94) & 1.23 (0.57) & 2.13 (1.25)\\
    & \xmark & \cmark & \cmark & \cmark & \textsc{P} & \textsc{C} & \textbf{1.62 (1.10)} & 2.04 (1.32) & \textbf{2.48 (1.22)} & \textbf{1.69 (1.20)} & \textbf{3.87 (1.46)}\\
    & \cmark & \cmark & \cmark & \cmark & \textsc{P} & \textsc{C} & 1.61 (1.10) & \textbf{2.06 (1.32)} & 2.43 (1.22) & 1.68 (1.22) & 3.79 (1.28)\\
    
    \bottomrule
    \end{tabular}}
\vskip -0.05in
    \caption{We compare the effect of different levels of context on prediction quality. We report the mean (standard deviation) on scores from 1--5 given by \textsc{GPT-4o} on our rubric.}
    \vskip -0.15in
  \label{tab:ablation}
\end{table*}

\squishlist
\item {\bf Semantic Similarity.} Predicted utterances must convey a similar  meaning as the original utterance.

\item {\bf Intent Preservation.} The response should serve the same conversational function, such as offering help, confirming information, or asking a question.

\item {\bf Specific Information Hallucination.} The model should avoid fabricating details; placeholders (e.g., ``xx:xx'' for a time) are acceptable, but concrete information should not be invented.

\item {\bf Contextual Appropriateness.}  The response should integrate smoothly into the surrounding conversation and maintain the natural flow.

\item {\bf Summary Alignment.} If both the original and predicted responses were summarized, the resulting summaries should be essentially equivalent.

\squishend
Together, these metrics provide a  framework for reconstructing turns that preserve meaning, function, and contextual coherence (see prompt in~\xref{apdx: eval prompt}).

{\it Precision-Recall metrics.} We ask the LLM evaluator to compute precision and recall by listing all important details in both predictions and ground truth, then counting overlaps to identify true positives, false positives, and false negatives. The evaluator is instructed to treat placeholders for unknown facts  as equivalent to ground truth to avoid penalizing anti-hallucination behavior. While not perfect, LLM-based detail extraction provides a reasonable, explainable approximation (see \xref{apdx: eval prompt}).

\begin{figure}[t!]
\centering
  \includegraphics[width=\columnwidth]{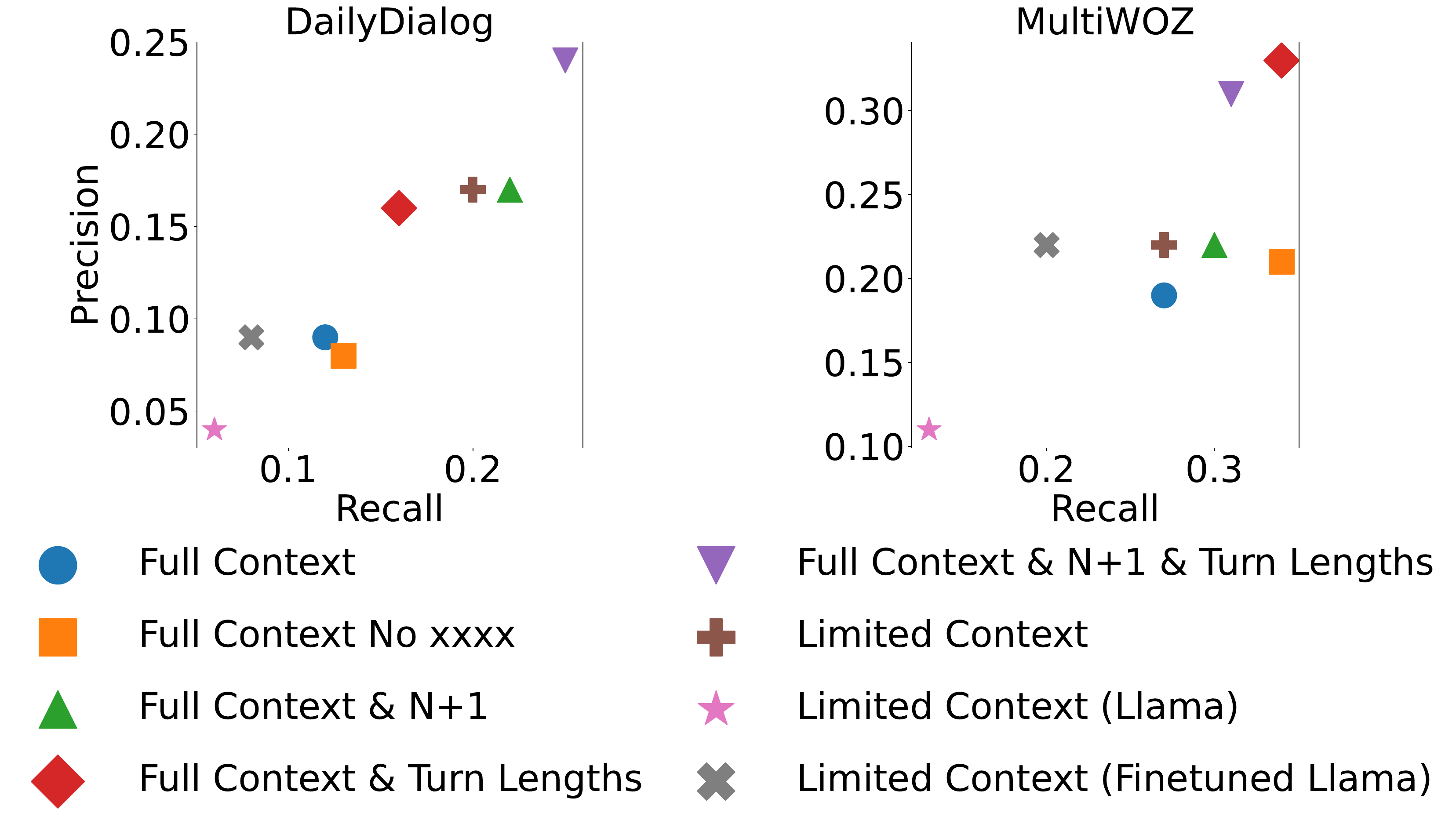}
  \vskip -0.05in
  \caption{Using our extraction based metrics, we show  macro-averaged precision and recall score for each dataset ($n=$ 493 for DailyDialog, $n=$ 705 for MultiWOZ). Note that as the details for the values are extracted by the evaluator LLM, the absolute numbers are not as meaningful as the relative differences between  methods. (Model is Claude unless specified.)}
  \label{fig:precrec}
  \vskip -0.15in
\end{figure}

\vskip 0.05in\noindent{\bf Results.} We evaluate each of our datasets on each contextual setting defined in \S \ref{sec:opr:task}. Rubric scores are shown in Table~\ref{tab:ablation}. For more structured conversations (such as those of DailyDialog and MultiWOZ), we find that the more context that we include --- both in user utterances and mask lengths --- the more accurate and highly rated the predictions are. We also find that even with finetuning, small models are unable to achieve the performance of larger models, even with the same level of local context.  Furthermore, in experiments where we remove the instruction to replace any specific information not available in the context with ``xxxx'', we find that performance is degraded across all measurements. We also investigate precision and recall (Fig.~\ref{fig:precrec}), 
confirming that additional context is generally better. \xref{apdx:oprex} provides an example of the model evaluation for the same conversation across multiple contexts.

Our results show that reconstruction does not support immediate turn-level intervention before the user's next utterance, but fits well into frameworks like LlamaPIE \cite{llamapie} and Memoro \cite{10.1145/3613904.3642450}, where assistance triggers after observing the user's response. Concretely, the agent observes turn t, reconstructs the likely interlocutor turn between t and t+1, then evaluates the user's actual response at t+1 against this inferred context, detecting omissions or misunderstandings (e.g., missed questions or constraints) and providing corrective guidance before the next turn. The agent only speaks when the user has missed something, minimizing interventions while maintaining low latency, all without requiring access to the true interlocutor utterance in real time.


\begin{figure}[t]
    \centering
    \subfloat[Example 1: Best Case]{%
      \includegraphics[width=0.9\columnwidth]{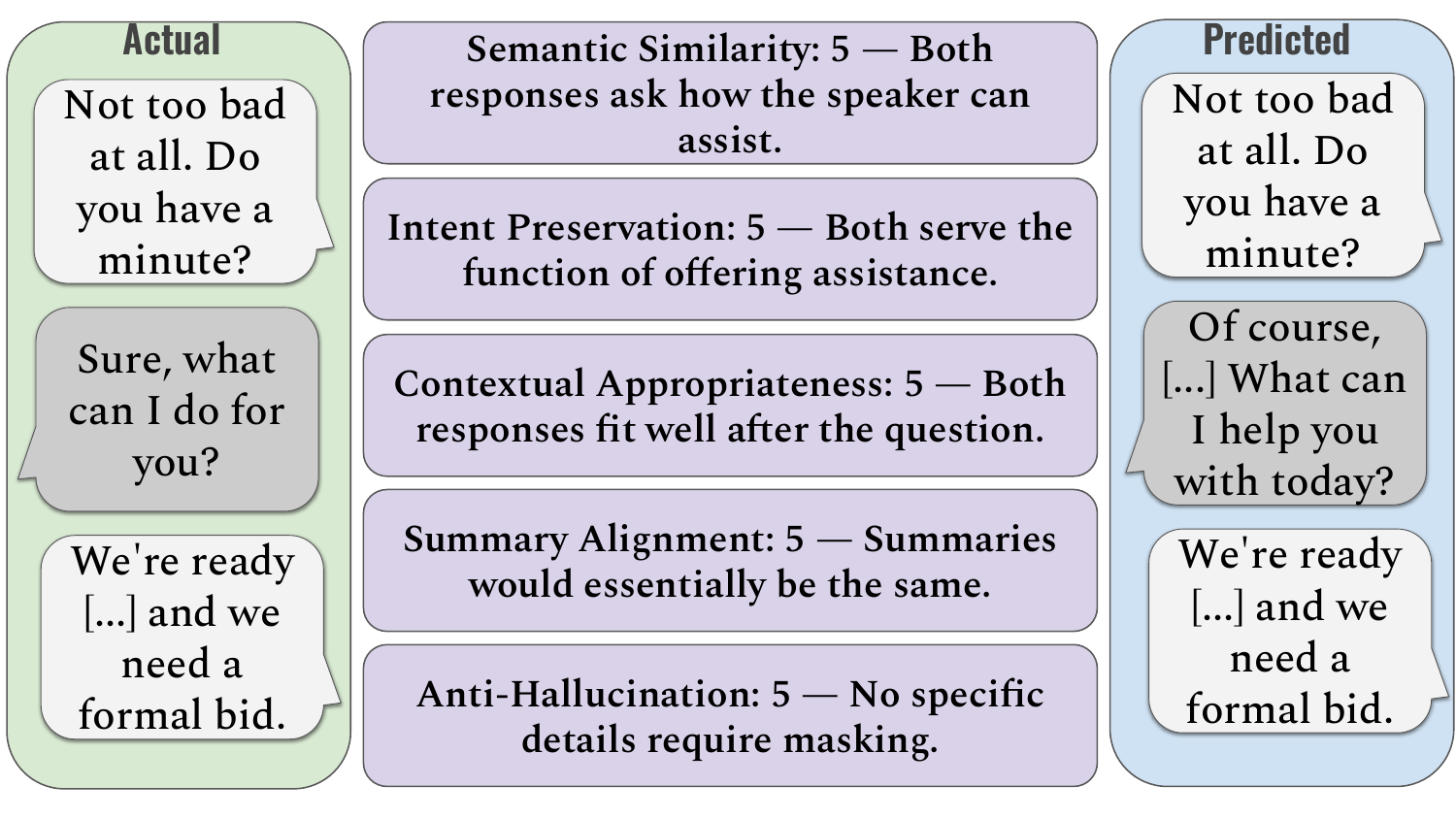}%
      \label{fig:examples:goodex}%
    }\qquad
    \subfloat[Example 2: Average Case]{%
      \includegraphics[width=0.9\columnwidth]{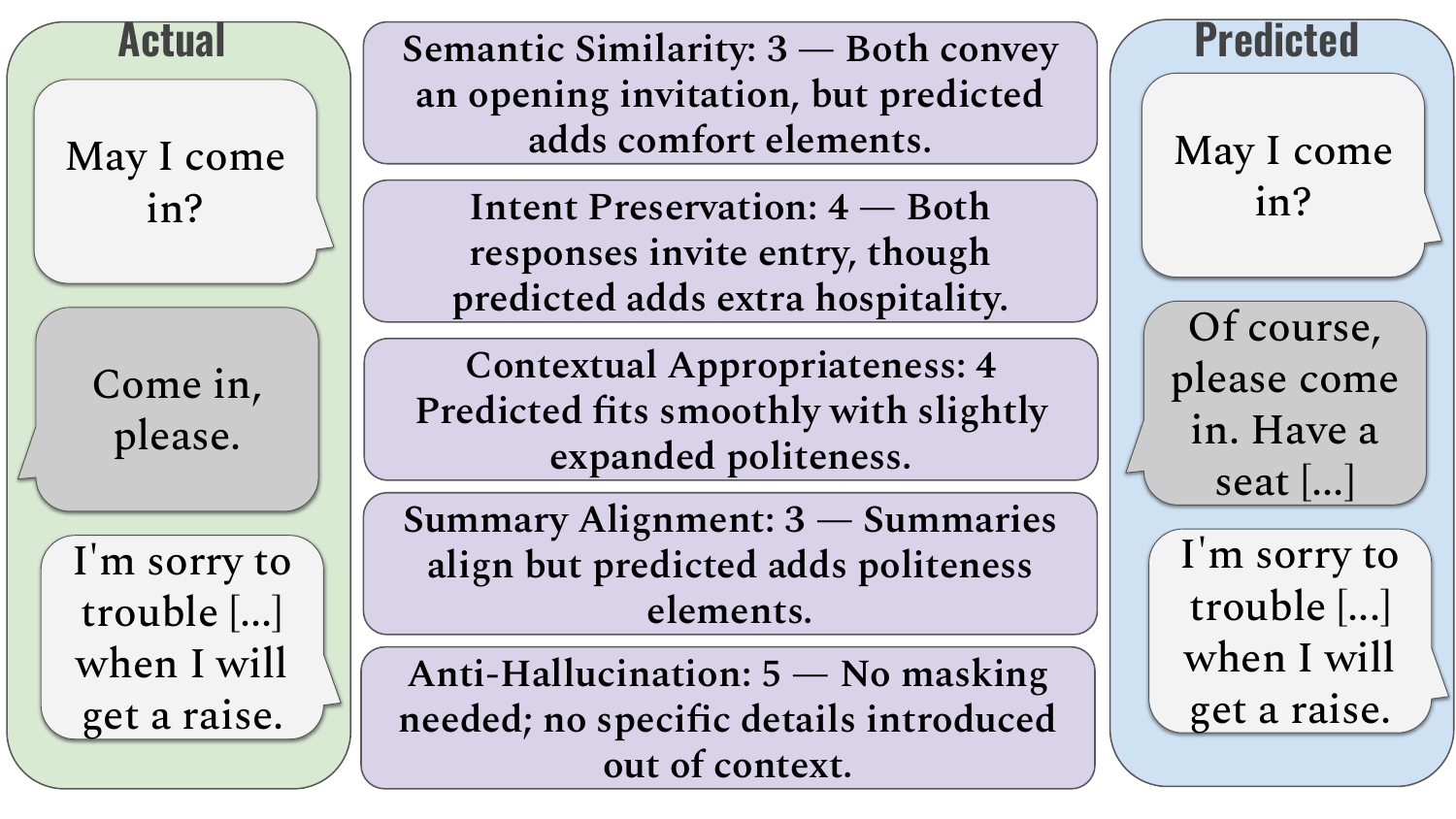}%
      \label{fig:examples:avgex}%
    }\qquad
    \subfloat[Example 3: Failure Case]{%
      \includegraphics[width=0.9\columnwidth]{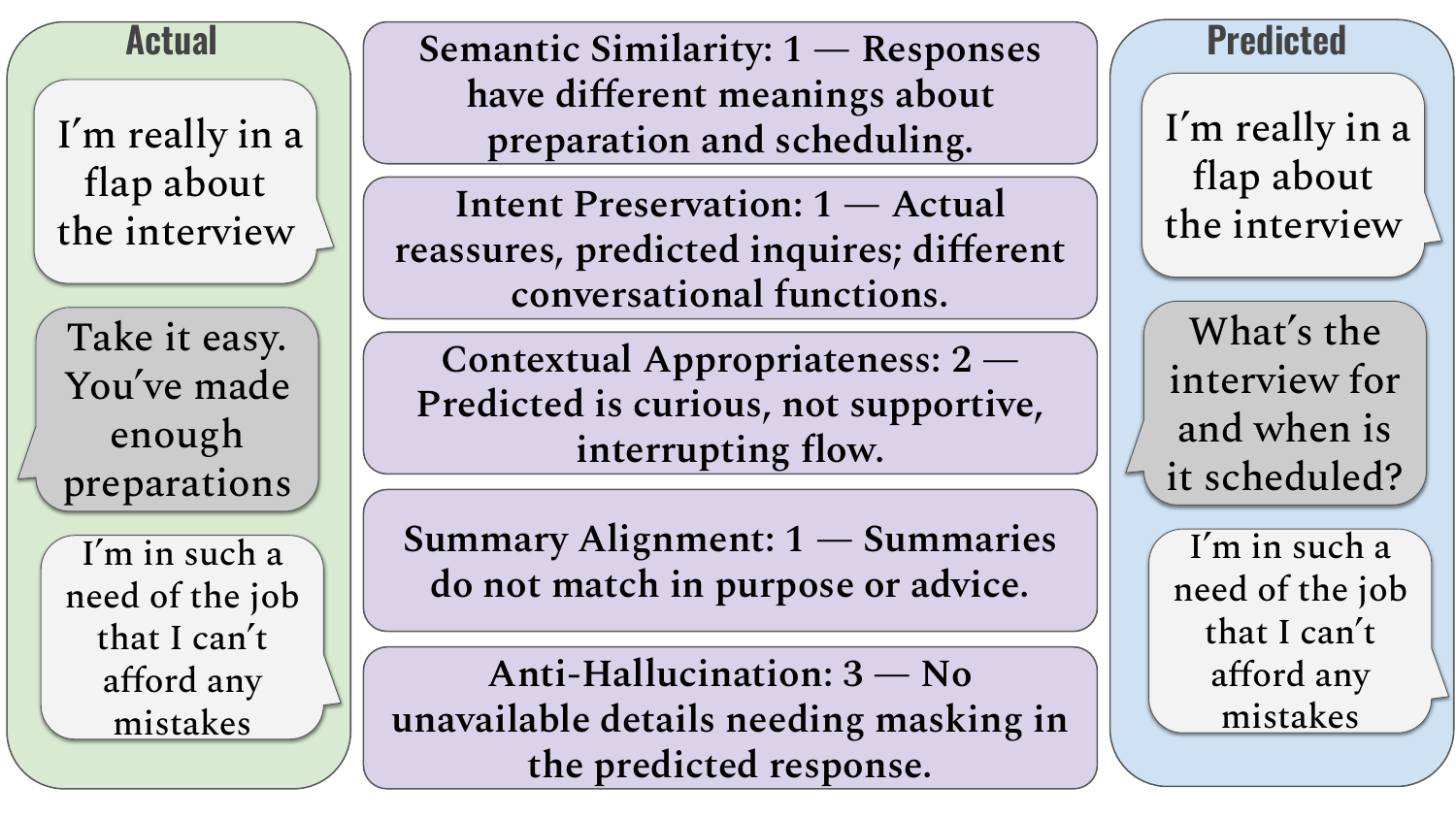}%
      \label{fig:examples:badex}%
    }
    \vskip -0.05in
    \caption{Example cases of our evaluation rubric for other party reconstruction showing high (a), average (b), and low (c) rubric scores. 
    }
    \vskip -0.15in
    \label{fig:predictionexamples}
\end{figure}

We also experiment with predicting the entire conversation at once rather than turn by turn. 
We reconstruct conversations across the full test splits for both DailyDialog and MultiWOZ and find the turn-by-turn predictions achieve higher scores across our rubric.
Results are shown in Table~\ref{tab:fullconvo} in \S \ref{sec:appen:multi}.

\section{1SC Summary Generation} \label{sec:summ}

We next turn to the task of summarizing a one-sided conversation, again with initial focus on the text-based datasets.

\subsection{Task Formulations}

We evaluate two variants: (a) summaries generated directly from one-sided transcripts with the masked speaker’s turns hidden, and (b) summaries built after reconstructing those turns using Turn N predictions from \S\ref{sec:opr}. For reconstruction, we use the overall best-performing setup, i.e., complete prior context with Turn N+1 and turn lengths generated with Claude. Comparing these variants reveals tradeoffs between reconstruction-free and reconstruction-heavy strategies for handling incomplete conversational data.

\subsection{Datasets and Methods}

We use the same datasets as in \S\ref{sec:opr:data}, using their test splits since no finetuning is performed for summarization. Unlike the recreation task, which has ground-truth Turn N responses, none of these datasets include summaries. Prior work shows Claude performs well at summarization~\cite{TheC3}; we therefore generate a full-conversation \emph{oracle} summary with Claude, by prompting it to create a  summary using the  \emph{two-sided} conversation, with the masked speaker's turns unmasked.   

Summaries from conversations with both masked and predicted turns are also generated by prompting Claude. Using the same model for both oracle and 1SC summaries provides a consistent and fair comparison.  The model is instructed to produce a comprehensive summary covering the conversation’s purpose, speakers’ goals, request flow, key details, and outcome. Placeholders (“xxxx”) are treated as the relevant obscured information. The full  prompt is provided in~\xref{apdx: summ create prompt}.

\subsection{Experiments}\label{sec:summ:exp}
For both our human and automated evaluation we use a 5 point rubric to examine  the summaries:
\squishlist
    \item \textbf{Content Coverage}: How well does the summary capture all the key specific information and main points from the original dialogue?
    \item \textbf{Dialogue Flow}: How well does the summary reflect the natural interaction between speakers?
    \item \textbf{Information Accuracy}: How accurate and faithful is the summary to the available information?
    \item \textbf{Purpose Outcome}: How clearly does the summary convey the dialogue's goals and results? 
    \item \textbf{Detail Balance}: How well does the summary balance important details from both speakers?
\squishend

To avoid unfair penalization, the instructions to both humans and the LLM-judge clarify that summaries should not be downgraded for using placeholders such as ``xxxx'' in place of specific details.  

\subsubsection{Human Evaluation}\label{sec:summ:humeval}
We conducted a human evaluation to evaluate conversation summary quality. Due to the long nature of summaries, this study was performed with a smaller number of human judges ($n=5$). Judges saw the full dialogue and three alternative summaries: one created directly from one-sided transcripts, one from reconstructed conversations using Turn N predictions, and one from the full, two-sided transcript. They rated each summary on a 1--5 scale for each of the rubric dimensions. Each session included an average of 20 comparisons, with the order randomized to control for position bias, and no labels provided.

We find that humans tended to perceive the dialogue flow and detail balance as better in the summary created from the predicted conversation than the masked conversation, but the information accuracy is predictably higher in the summary from the masked conversation.  This is especially true of DailyDialog as compared to MultiWOZ. Detailed results are shown in Fig.~\ref{fig:humsummrubric} in \S\ref{apdx: figs}.

\subsubsection{Automated Evaluations}\label{sec:summ:autoeval}

\textbf{Metrics.} As before, we use both rubric scores and precision-recall metrics. 

{\it Rubric scores.} As in the human evalution, we adopt a blind review setup. The evaluator model (\textsc{GPT-4o})  is presented with three summaries, randomly labeled A, B, or C. Alongside the summaries, the full conversation is provided for reference. The model is instructed to score each summary across each of the five rubric dimensions using a 1–5 scale.
In addition to assigning scores, the evaluator is also asked to produce a relative ranking of the three summaries from best to worst. The full evaluation prompt is included in \S\ref{apdx:blindevalprompt}.

\begin{figure}[t!]
    \centering
    \begin{subfigure}[t]{\columnwidth}
        \centering
        \includegraphics[width=\columnwidth]{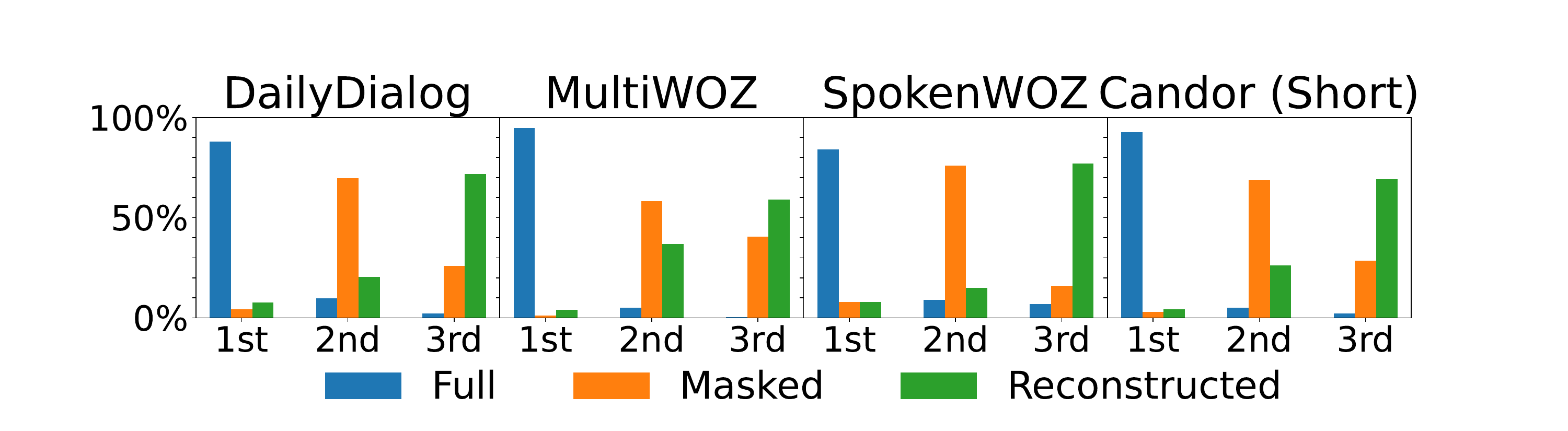}
        \caption{Summary ranking results from our LLM Judge.}
        \label{fig:summrank}
    \end{subfigure}
    
    \vskip 0.05in  

    \begin{subfigure}[t]{\columnwidth}
        \centering
        \includegraphics[width=\columnwidth]{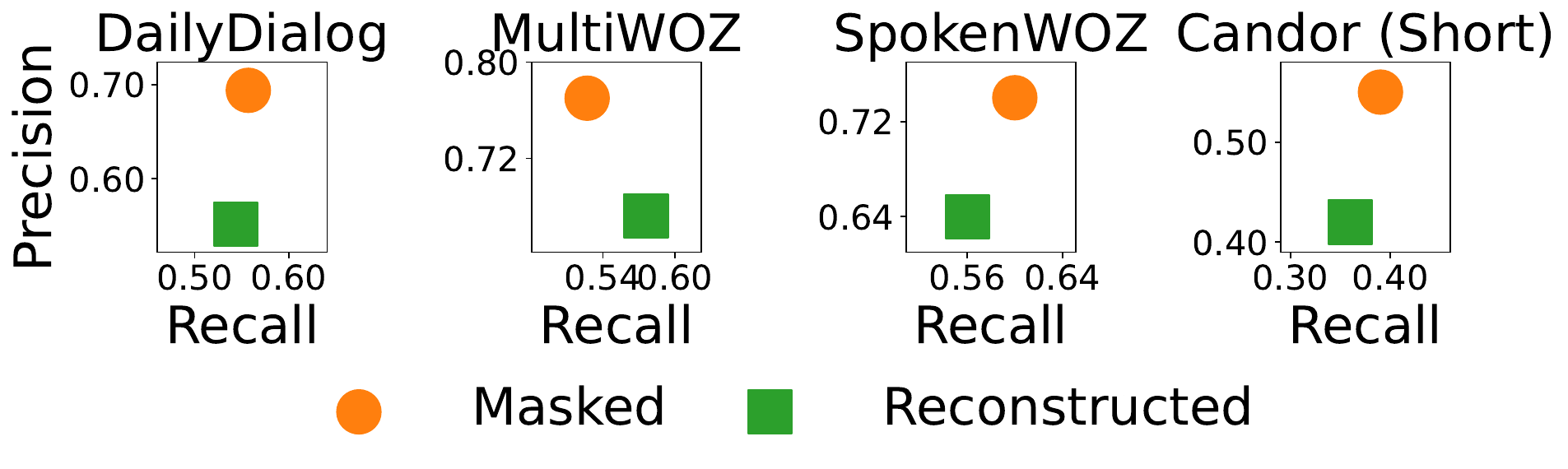}
        \caption{Macro-averaged precision and recall results.}
        \label{fig:sumprecrec}
    \end{subfigure}
    \vskip -0.05in
    \caption{Results for summary evaluation. (DailyDialog $n=1000$, MultiWOZ $n=1313$, SpokenWOZ $n= 100$, Candor $n=137$). Masked-dialogue summaries were consistently ranked above predicted-dialogue summaries, and had higher precision and comparable recall.} \label{fig:summary_results}
    \vskip -0.15in
\end{figure}

\textit{Precision-Recall metrics.} The evaluator model is also asked to compute precision and recall values for the masked and predicted summaries, using the oracle as a baseline, following \S \ref{sec:opr:eval:llm}. We provide the informed evaluator prompt in \xref{apdx:informedevalprompt}.

\vskip 0.05in\noindent{\bf Results.} Our automatic and human evaluations broadly agree: fully masked dialogues produce just as good if not better summaries than their reconstructed counterparts, but reconstruction may lend a hand in more task oriented settings. As expected, the oracle summaries from our full (two-sided) conversations perform the best and have the most consistent ratings. The two one-sided summaries are very close in scores, with the fully masked conversations slightly outperforming the predicted on average, particularly for less task-oriented utterances. For full details see Fig.~\ref{fig:summrubric} in \xref{apdx: figs}. 

Considering the ranked orders of the summaries by the evaluator model further confirms this finding.  Fig.~\ref{fig:summrank} shows a detailed breakdown of the summary rankings. From these datasets, we find that the reconstruction-free summaries were preferred to the reconstruction-heavy summaries. Across datasets, our masked-dialogue summaries also have consistently higher precision than the reconstructed-dialogue summaries. Recall on masked-dialogue summaries is better in conversational settings, but in the task-oriented dialogues of MultiWOZ reconstructed-dialogue summaries have higher recall. Full details in  Fig.~\ref{fig:sumprecrec}. 

Despite our finding that reconstructed utterances  convey intent similar to the ground truth, summaries based on these reconstructed utterances performed worse than those without them. This apparent contradiction can be explained by considering the goals of each task. In other-party reconstruction, the objective is to generate plausible utterances given the local context, with less emphasis on factual correctness. Across all ablations, semantic similarity to the ground truth scored lower than contextual appropriateness within the conversation.

In contrast, summary generation aims to produce faithful representations of the conversation as a whole. Plausible reconstructions may introduce details that were not present in the original dialogue; when such details propagate into the summary, they reduce factual alignment with the ground truth. Masked inputs avoid introducing incorrect information, so although they omit content, they yield summaries that are more factually accurate. Thus, reconstructed turns may be locally preferred for plausibility while still resulting in worse summaries under a faithfulness-based evaluation.


\section{Real Dialogue Evaluation}\label{sec:real}
Our experiments with DailyDialog and MultiWOZ point to a need for expanded context in other party reconstruction (\S\ref{sec:opr}) and  no need for reconstruction for summary generation (\S\ref{sec:summ}). To determine whether this pattern holds for more realistic settings, we evaluate both reconstruction and summarization using Candor and SpokenWOZ; we use the transcripts from these audio based datasets. While care was taken in the development of SpokenWOZ to ensure a lack of overlaps and backchannels between the speakers, Candor has raw, unscripted, unadulterated dialogues from video chat correspondences. We preprocess Candor to exclude anything that was said by the second speaker (our masked speaker) during the duration of our user's utterances. This preprocessing mimics the one-sided setting where we would only know that someone else had spoken by the silence in the recording of the user and simulates the true lengths of spaces between user utterances that would be captured in a one-sided recording setup, thus representing the most realistic scenario.

\subsection{Other Party Recreation} \label{sec:real:recreate} 
With both spoken datasets, we test the five full-prior context settings\footnote{Preliminary experiments with limited context and Llama as well as the results from the text-based dataset experiments showed substantial decreases in the rubric scores for these  settings. We leave further experimentation to future work.} and evaluate using both human annotators and LLM-as-a-judge methods. 

For the human evaluation
we create three-turn examples from the reconstructions using Turn N predictions with turn lengths and Turn N+1 given --- the same reconstructions used for summary generation in \S\ref{sec:summ}. The annotation process is the same as detailed in \S\ref{sec:opr:eval:hum}.

Our results show that on SpokenWOZ raters see no difference in the quality of the response between the real ground-truth utterance and the reconstructed utterance. With Candor, our reconstructed utterances were often preferred over those of the true conversation. The lower half of Table~\ref{tab:humanrecon} gives details on the human judgements. These results are consistent with those of the text-based conversations:
our model produces reasonable utterances given a limited context.


The LLM-as-a-judge results are shown in the lower half of Table~\ref{tab:ablation}.  Comparing SpokenWOZ to MultiWOZ,  working with audio transcripts leads to lower performance.  Comparing SpokenWOZ to Candor,  task-oriented dialogue were easier to recreate than conversational dialogue. 

\subsection{1SC Summary Generation}
\label{sec:real:summ} 
We also evaluate the two spoken datasets on summary generation as described in \S\ref{sec:summ:exp}. In this case, we also construct summaries for smaller subsets of the Candor conversations, where each conversation is split into several smaller conversations of approximately 25 turns. We perform this split to allow for human evaluations of Candor summaries; human evaluation of long conversations is a prohibitively difficult task \cite{krishna-etal-2023-longeval}. 

Our human evaluations show similar trends to the text-based dialogues. Full results are in Fig.~\ref{fig:humanspokensummrubric} in \S\ref{apdx: figs}. The scores for the masked and predicted summaries are similar, with the largest gap in detail balance; summaries from masked conversations tend to be lacking in balance between the two speakers as there is no second speaker to extract details from. As with recreation (\S\ref{sec:real:recreate}), summarization of one-sided conversations from Candor were more difficult than from SpokenWOZ. The average two-sided oracle summary ranking remains consistent, but the scores for the one-sided conversation types drop. While our human evaluation of MultiWOZ showed that reconstruction may be helpful in a task-based environment, the realistic conversations from SpokenWOZ do not follow the same trend.

The LLM-as-a-judge results are shown in Fig.~\ref{fig:spokensummrubric} in \S\ref{apdx: figs}. The most interesting finding of this experiment is that when evaluating the summaries for the entire Candor dialogues, the results were contrary to every other dataset: predicting the missing utterances was helpful in constructing good summaries. 

We conjecture that this difference is caused by the length of Candor's dialogues (at least ten times longer than the other datasets, on average). To remove this factor, we evaluate on the subset of Candor dialogue excerpts that we used for human evaluations.  
Comparing this setting to SpokenWOZ and DailyDialog, the results revert back to our original findings: reconstruction provided worse summaries than simply leaving the utterances masked. Relative rankings and precision/recall scores are shown in Fig.~\ref{fig:summary_results}. We also find that unlike the reconstructed turns from \S\ref{sec:real:recreate}, the difference in scores between the text-based dialogues and the spoken dialogue transcripts are minimal: working with audio transcripts results in little to no drop in performance. 


While other party reconstruction on transcripts of spoken conversations is not at the level of text-based conversations, summarization in this real-world case is viable under our current settings.

\section{Conclusion}
We introduce the one-sided conversation (1SC) problem and study two tasks: reconstructing the missing speaker’s turns and generating faithful summaries from one-sided input. Our work provides a foundation for this problem and opens future directions, including multi-turn prediction and leveraging past predictions in online settings.

The 1SC problem has broad implications for privacy-aware conversational AI in telemedicine, call centers, and personal assistants, where capturing only one side of a dialogue respects privacy while enabling downstream support such as proactive guidance, contextual memory, and efficient documentation. By formalizing 1SC, we move toward systems that reason effectively under asymmetric dialogue while aligning with real-world legal and social constraints.

Our introduction of the 1SC problem opens up many areas of research. Future directions include  post-training (e.g., RLHF,  RLAIF) specifically to improve the reconstruction and summarization of 1SCs, using past predictions as context for future predictions (e.g having the model predict turn 2, then use that prediction and the 1SC as context for predicting turn 4). Further work  could also tackle conversations with more than two parties.

\section*{Limitations and Risks}
{\bf Limitations.} While promising, our work has several limitations which open up opportunities for future research. First, reconstruction inevitably introduces uncertainty: even with placeholder prompting, models may still hallucinate details or subtly shift intent. Second, there remains a performance gap between large and smaller models on the 1SC task, despite finetuning the smaller models; further investment in finetuning data or methods  may  close this gap. Third, unlike structured applications such as call centers or telemedicine, free-flowing conversations, such as those in the Candor corpus, feature frequent and abrupt topic shifts, which make the tasks more challenging. Fourth, the use of turn N+1 requires slightly delayed inference capabilities, meaning that applications like proactive assistants must be designed with this in mind. Finally, in many settings, the use of a third-party LM API, even with only one side of a conversation, may pose a privacy risk; a local deployment of an open-weight model may be preferable.

We also note that with closed models, there is always a risk of data contamination, where the training data of the model may have contained our datasets.  We highlight three key points: 1)   Candor, the real-world conversation dataset used in our evaluation, was likely not used for training.  2) Dataset contamination would likely be an issue where scores (especially ``Semantic Similarity'') are very high on reconstruction.  This was not the case in our experiments, which  serve to show the limitations of even large models (even with possible contamination). 3) Contamination is a widely faced problem in NLP right now, as the training data for most language models is undisclosed. Even for open models, there is no easy or perfect way to quickly determine what datasets exist in the training data for models (this is essentially the same as the widely studied decontamination problem).  Even if we were to do an exact match on training data for open models, the data may exist within the model in a slightly altered form, so a more sophisticated fuzzy match may be needed. For now, the accepted convention in recent NLP papers has been to assume contamination is not a factor in experiments with these specific datasets; see \citet{zhu-etal-2024-large}, who benchmark LLaMa on MultiWoz, \citet{10.1145/3583780.3614949},  who evaluate LLaMa on DailyDialog, and \citet{shin-etal-2025-eco}, who evaluate MultiWOZ and DailyDialog on LLaMa.

\noindent{\bf Ethical Considerations.} 
While we frame 1SC as a step toward privacy-conscious AI, summarizing or reconstructing one-sided conversations can still raise concerns around sensitive data, particularly if speakers mistakenly assume their conversations are private. Indicators such as a visible LED on smart glasses could signal to others that a device is operating in 1SC mode. Additionally, clearly labeling summaries or predicted turns as AI-generated rather than authentic transcripts is critical for setting correct user expectations and accurately conveying the outputs.

It is important to acknowledge that highly accurate predictive models in the future could nearly replicate the exact words spoken in a conversation. Therefore, future deployments must pair technical advances with robust safeguards, including transparent communication of model uncertainty and strong protections for the storage, use, and sharing of derived summaries. Addressing these risks will be essential for deploying 1SC systems in socially and legally responsible ways.

\section*{Acknowledgments}
We thank Jacob Edelson and Nicholas Batchelder  for their help in early iterations of this project, particularly with initial code outlines. 

\bibliography{custom}


\appendix

\section{Further Related Works} \label{sec:appen:related}
\noindent{\bf Text and code infilling.} 
Our work on reconstructing missing speaker turns builds on text infilling, where models generate spans absent from a sequence. Masked language models like BERT~\cite{devlin-etal-2019-bert} established span-level prediction, later extended to arbitrary-length infilling~\cite{donahue-etal-2020-enabling,bavarian2022efficienttraininglanguagemodels}. Other efforts edit incomplete utterances~\cite{pan-etal-2019-improving,jiang1,du-etal-2023-multi,li-etal-2025-dialogue}, but these focus on rewriting rather than filling missing dialogue. In contrast, we are the first to address the one-sided conversation setting, where all turns from one speaker are absent.

Parallel work in code infilling trains models to complete missing code blocks~\cite{chen2021evaluatinglargelanguagemodels,wang-etal-2021-codet5,fried2023incodergenerativemodelcode,10.1145/3695988}, requiring local fluency and global consistency. This structural analogy underscores our task:  code infilling must preserve syntax and semantics, while dialogue infilling must maintain turn-taking, intent, and conversational flow.

\vskip 0.05in\noindent{\bf Dialogue state tracking and user simulation.} Dialogue state tracking~\cite{williams-etal-2013-dialog,10.5555/3495724.3497418,king-flanigan-2023-diverse} maintains representations of user goals and intents across turns. User simulators~\cite{schatzmann-etal-2007-statistical,crook2017sequence,wan-etal-2022-unified,sekulic-etal-2024-reliable} generate plausible missing utterances in task-oriented dialogue, offering controlled training environments for conversational agents. These approaches resemble our goal of reconstructing missing turns but typically rely on structured annotations (slots, schemas, intents), domain constraints (task-oriented settings), and partial or full access to both speakers’ turns or goals.

\vskip 0.05in\noindent{\bf Speech and one-sided recordings.} Tasks like speaker diarization~\cite{Hrknen2024EENDM2FMM,10446213} and speech separation~\cite{10.1145/3613904.3642057,itani2025tfmlpnettinyrealtimeneural} handle overlapping speakers but focus on signal-level recovery, not generating missing text. Sociolinguistics and privacy research study one-sided or overheard recordings in terms of awareness and privacy-preserving features~\cite{privacy1,10.5555/3692070.3694441,POHLHAUSEN2026101823}. In call centers and telemedicine, only one side’s audio or transcript is typically retained, yet, to our knowledge, no prior work attempts to reconstruct the missing turn text. We frame this problem as text generation under one-sided observability.

\vskip 0.05in\noindent{\bf Conversational assistance.} Recent work enhances human interactions with real-time guidance in call centers and medical consultations~\cite{llamapie,zhu2025overhearingllmagentssurvey,proactiveagent-emnlp}, but assumes access to both sides of the dialogue and ignores recording constraints. In contrast, we explicitly account for these limits, showing that inferring missing turns can still enable the streaming applications envisioned by prior work.

\section{Further Motivation}\label{sec:appen:motiv}
Real-time proactive assistants can support neurodiverse individuals, by helping them interpret social cues~\cite{10.1145/3654777.3676430}; act as memory augmentation tools by providing contextually relevant reminders during conversations~\cite{10.1145/3613904.3642450, 10.1145/3394171.3413663}; and enhance professional interactions including sales calls and negotiations~\cite{10.1145/3626772.3657843}, interviews~\cite{7579163}, customer-care interactions~\cite{10.1145/3626772.3657843}, and cross-cultural communication~\cite{10.1145/3196709.3196734}. 

Memory augmentation and proactive conversational support have long been central interests in the HCI community~\cite{Myers2007ProactiveBO, 10.1145/3432193, BERUBE2024100411}. Since Vannevar Bush’s 1945 vision of the Memex~\cite{Bush1945}, 
researchers have proposed numerous prototypes, from lifelogging systems that continuously record user behavior to just-in-time retrieval systems that surface relevant information based on situational context. However, despite this rich design history, the NLP community has only recently begun to tackle the practical challenges required to make these systems viable in real-world use.

\section{Data Details}\label{sec:appen:data}
MultiWOZ and SODA have predefined train/test splits. For DailyDialog, we split 80/10/10 for train/validation/test. For both SpokenWOZ and Candor, the additional length from the nuances of spoken dialogue lead to longer conversations; given  our budget constraints, we  downsample to create test sets  and make predictions on  just over 2000 masked turns each (equating to 100 SpokenWOZ conversations and 8 Candor conversations). In all datasets, the first speaker is designated the user and the second the masked speaker. Table~\ref{tab:splits} breaks down our data and gives details on the number of masked turns per dataset

\begin{table}
    \centering
    {\footnotesize

    \begin{tabular}{lrrrr}
        \toprule
        \multirow{2}{*}{\textbf{Dataset}} & \multirow{2}{*}{\textbf{Train}} & \multirow{2}{*}{\textbf{Val}} & \multirow{2}{*}{\textbf{Test}} & \multirow{2}{1cm}{\textbf{Average Length}}\\
        & & & & \\
        \midrule
        MultiWOZ & 61k & 6.4k & 6.4k & 24 turns\\
        \midrule
        DailyDialog & 32.3k & 3.9k & 4.5k & 8 turns\\
        \midrule
        Candor & -- & -- & 2.0k & 426 turns\\
        \midrule
        SpokenWOZ & -- & -- & 2.0k & 41 turns\\
        \midrule
        SODA & 3.5M & 436k & -- & 8 turns\\
        \midrule
        \textbf{Total} & 3.6M & 446k & 15k & --\\
        \bottomrule

    \end{tabular}
    }
    \vskip -0.05in
    \caption{Number of masked turns in each dataset.} 
    \vskip -0.15in
    \label{tab:splits}
\end{table}

\section{Finetuning Details}\label{sec:appen:finetune}

The models are finetuned on the train splits of MultiWOZ, DailyDialog, and SODA. Of our combined $3,603,573$ train examples, $3,560,567$  were successfully processed into training data for \textsc{LLaMA-3.2-1B}. We trained each model using 1 GPU and stopped after approximately 3 days. We use the code provided by \cite{donahue-etal-2020-enabling}, and our usage is consistent with their intended use. All other finetuning details including learning rate and scheduler follow from those defined by \cite{donahue-etal-2020-enabling} in their paper.

\section{Annotator Details} \label{apdx:annotators}
\subsection{Demographics}
The study was approved by IRB. All participants were unpaid volunteers, provided consent, and were recruited from our institution and nearby areas. Respondents included 3 undergraduate students, 3 post-graduate industry professional, 9 PhD students and 1 Post-Doctoral Researchers. 75\% (12/16) of respondents were men, and the other 25\% (4/16) were women. Participants were informed that their rankings would be used in a research paper. 

\subsection{Data Selection}
We randomly sampled 100 limited-context dialogues:
50 from Claude predictions and 50 from our fine-tuned Llama model. The dialogues were split evenly between MultiWOZ and DailyDialog; 25 of each for each model. Each dialogue was reviewed by at least 6 of our 16 annotators. Full annotator instructions are provided in~\xref{appdx:abinst}.

\subsection{Instructions for A/B Test} \label{appdx:abinst}
Hello! Welcome to the AB testing script for One-Sided Conversations. Please follow the instructions to complete the AB testing task. You will be presented with pairs of responses for the same dialogue context. For each pair, please choose which response fits better in the context (1 or 2). If you feel both responses are equally good, press 0. You will not be penalized for choosing 0, but please use it sparingly. Some responses include XXXXXX rather than specific names, places, or numbers. Please treat these as normal words in the conversation, as if they were names, places or numbers. Thank you for your participation! 


    
    
    
    



\section{Multi-turn Prediction}\label{sec:appen:multi}

\begin{table*}[t!]
    \centering
    \small
    {
    \begin{tabular}{lcccccc}
    \toprule
    Dataset & Scenario
    & \makecell{Seman.\\Sim. ($\uparrow$)} & \makecell{Intent\\Pres. ($\uparrow$)} 
    & \makecell{Context.\\Approp.($\uparrow$)} & \makecell{Summ.\\Align.($\uparrow$)} 
    & \makecell{Anti-\\Halluc.($\uparrow$)} \\
    \midrule
    
    \multirow{2}{*}{\makecell{Daily\\Dialog}} 
    & Turn-by-turn & \textbf{3.56 (1.11)} & \textbf{4.21 (0.94)} & \textbf{3.96 (0.97)} & \textbf{3.30 (1.14)} & \textbf{4.77 (0.71)}\\
    & All at Once & 2.43 (1.27) & 3.27 (1.50) & 3.02 (1.28) & 2.45 (1.30) & 3.12 (1.36)\\
    
    \midrule
    
    \multirow{2}{*}{\makecell{Multi\\WOZ}}
    & Turn-by-Turn & \textbf{3.53 (0.74)} & \textbf{4.80 (0.43)} & \textbf{4.37 (0.68)} & \textbf{3.50 (0.78)} & \textbf{4.83 (0.55)}\\
    & All-at-Once & 1.29 (0.79) & 2.16 (1.32) & 3.14 (1.26) & 1.33 (0.84) & 3.25 (1.44)\\
    
    \bottomrule
    \end{tabular}}
    \vskip -0.05in
    \caption{Full conversation evaluation results. Performance is better when reconstructions are generated turn-by-turn. Values are reported as mean (standard deviation).}
    \vskip -0.15in
    \label{tab:fullconvo}
\end{table*}

Table~\ref{tab:fullconvo} shows full conversation evaluation results. Predicting turn by turn with our maximum tested context produced better results than predicting the entire conversation at once.

\section{Additional Experiments} \label{apdx:exp}
We also conducted a human A/B test of the two one-sided summary options (reconstruction-free or reconstruction-heavy) for DailyDialog and MultiWOZ in addition to the rubric evaluation of the summaries. We found that the summary from the masked conversation was preferred over the reconstructed conversation summary for DailyDialog, but that for MultiWOZ the reconstructed-conversation won out by a small margin.

\begin{figure}[h]
  \includegraphics[width=\columnwidth]{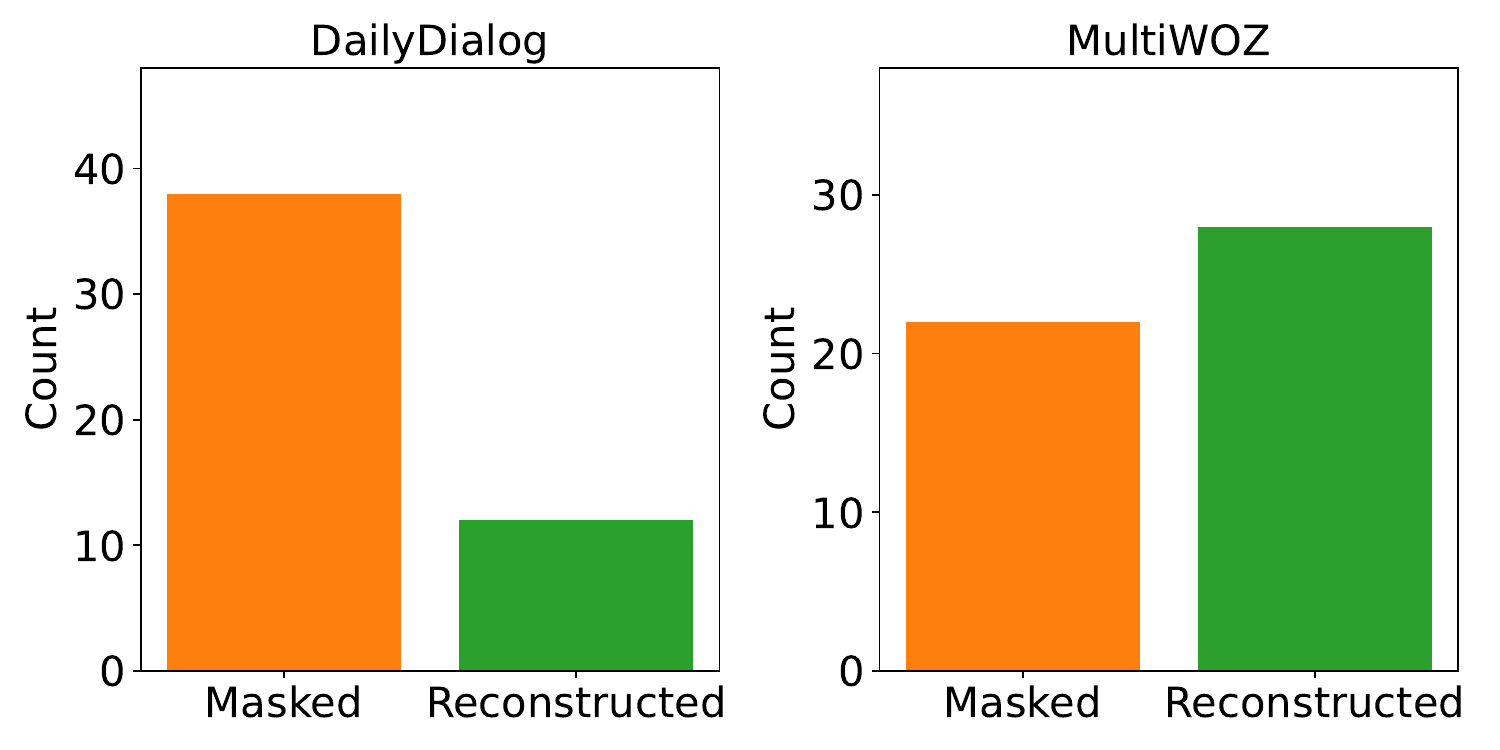}
  \caption{Ranking of DailyDialog and MultWOZ Summaries by human annotators. }
  \label{fig:sumhumrank}
\end{figure}

\begin{figure*}[h]
  \includegraphics[width=\textwidth]{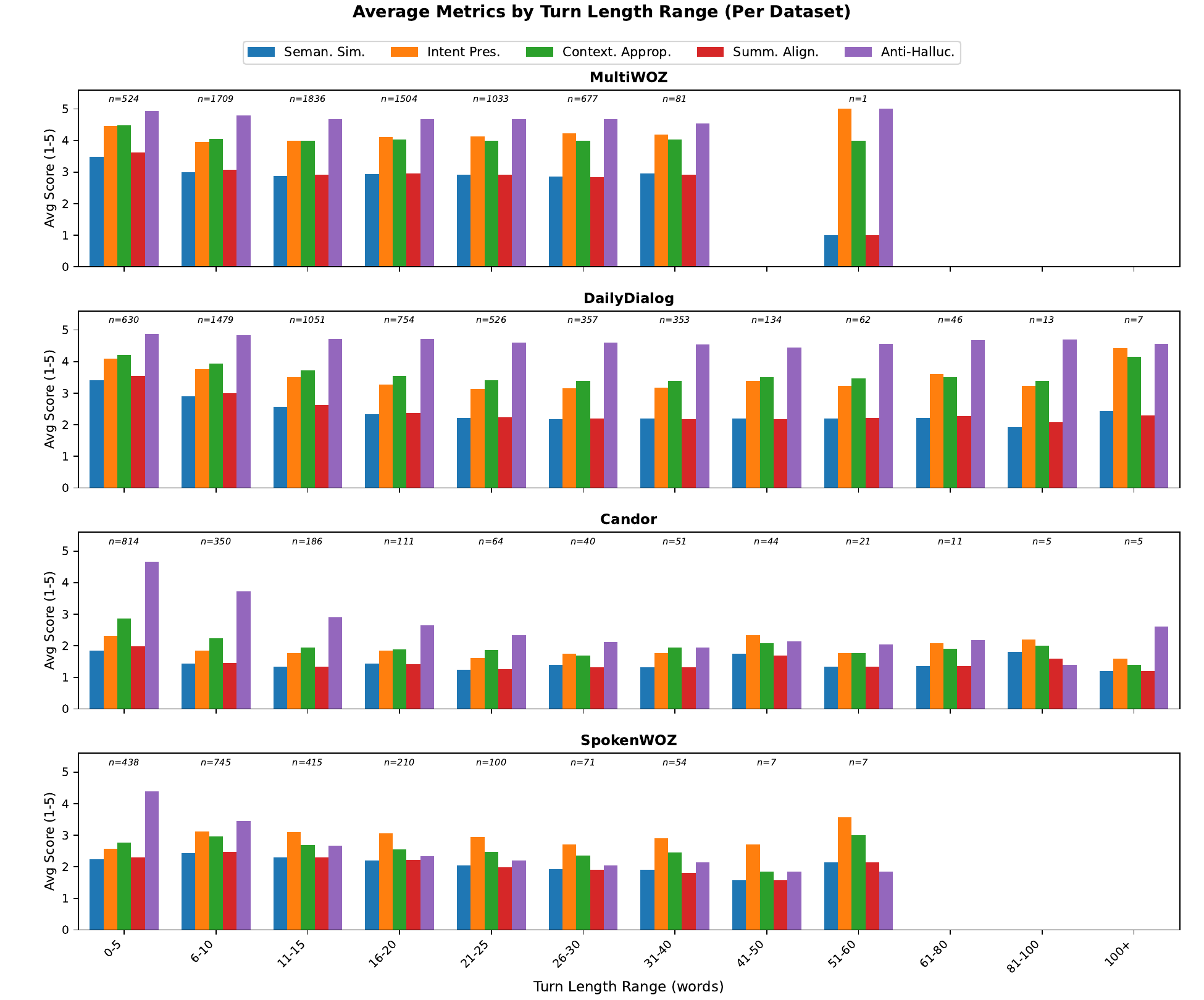}
  \caption{Average across reconstruction metrics on our best performing scenario distributed across turn lengths.}
  \label{fig:metricsbyturn}
\end{figure*}

We evaluated the effect of turn length on the average scores from our LLM-Judge on our text-based datasets for reconstruction accuracy, shown in Fig~\ref{fig:metricsbyturn}. We found no significant difference between scores for short and long utterances.

We evaluated the effect of masking the second speaker instead of the first. We tested both text-based datasets and evaluated using our LLM-Judge. Table~\ref{tab:speaker_masking_results} shows the results; we found no significant difference in scores depending on which speaker was masked.

\begin{table}
\centering
\footnotesize
\resizebox{\columnwidth}{!}{%
\begin{tabular}{lccccc}
\toprule
 \makecell{Masked\\Speaker} & \makecell{Seman.\\Sim. ($\uparrow$)} & \makecell{Intent\\Pres. ($\uparrow$)} & \makecell{Context.\\Approp.($\uparrow$)} & \makecell{Summ.\\Align.($\uparrow$)} & \makecell{Anti-\\Halluc.($\uparrow$)} \\
\midrule
\makecell{First\\(as in paper)} &
2.96 (1.20) &
4.06 (1.17) &
4.07 (0.98) &
3.00 (1.24) &
4.74 (0.75) \\
\midrule
\makecell{Second} &
2.98 (1.43) &
3.89 (1.27) &
3.74 (1.24) &
3.01 (1.45) &
4.64 (0.96) \\
\bottomrule
\end{tabular}%
}
\caption{Evaluation results under different speaker-masking settings. Values are reported as mean (standard deviation).}
\label{tab:speaker_masking_results}
\end{table}

\section{Additional Figures} \label{apdx: figs}
\begin{figure*}[h]
  \includegraphics[width=\textwidth]{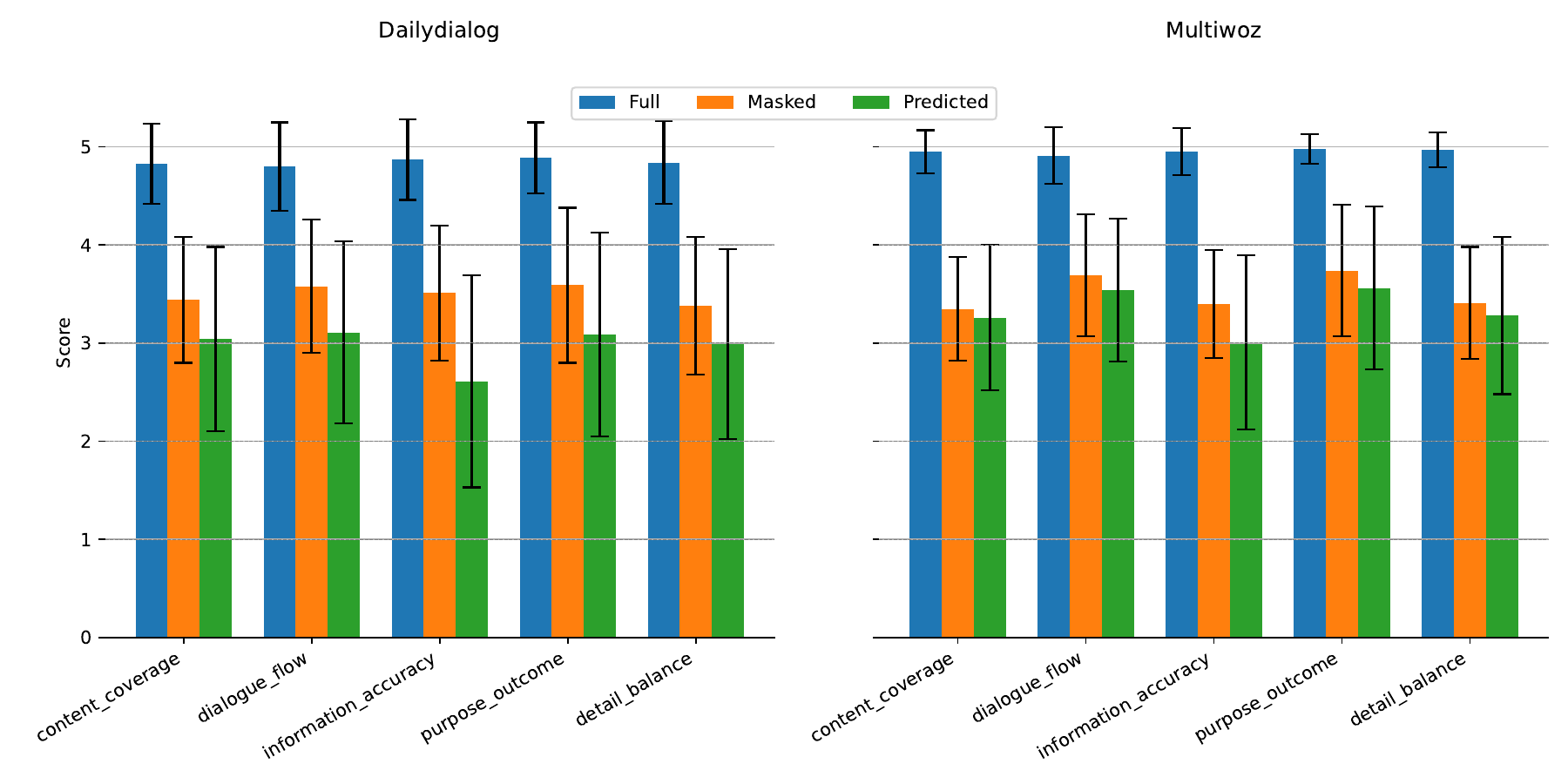}
  \caption{The results of our rubric evaluation on DailyDialog and MultiWOZ.}
  \label{fig:summrubric}
\end{figure*}

\begin{figure*}[h]
  \includegraphics[width=\textwidth]{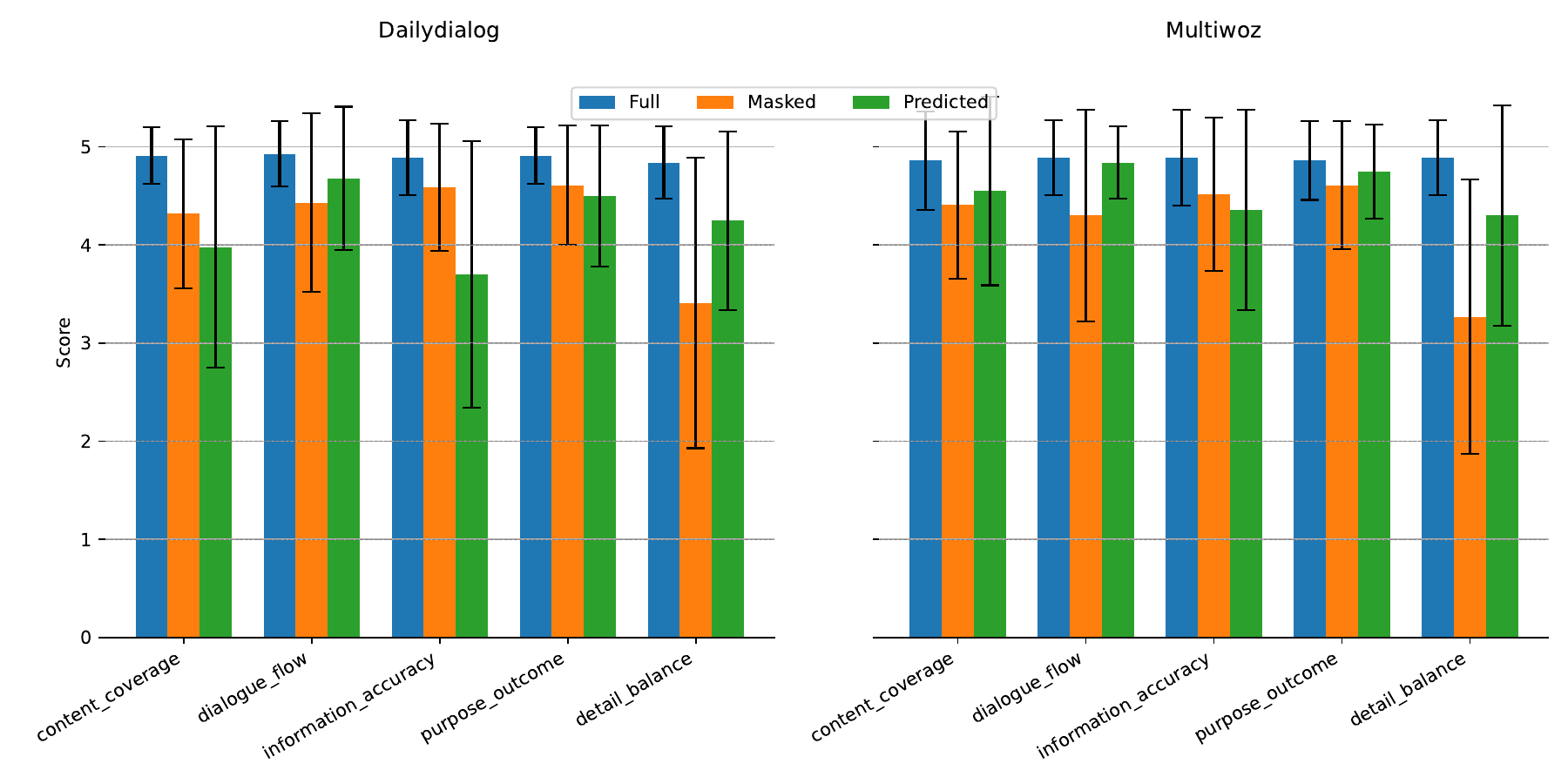}
  \caption{The results of our human rubric evaluation on DailyDialog and MultiWOZ.}
  \label{fig:humsummrubric}
\end{figure*}

\begin{figure*}[h]
  \includegraphics[width=\textwidth]{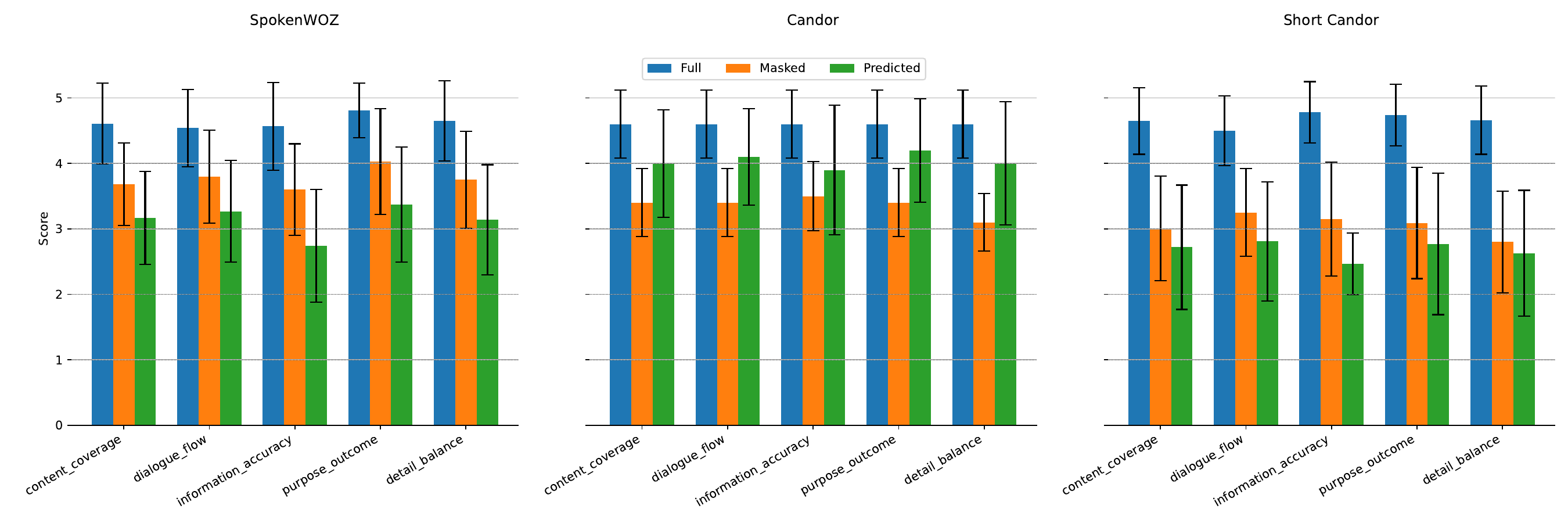}
  \caption{The results of our rubric evaluation on SpokenWOZ and Candor.}
  \label{fig:spokensummrubric}
\end{figure*}

\begin{figure*}[h]
  \includegraphics[width=\textwidth]{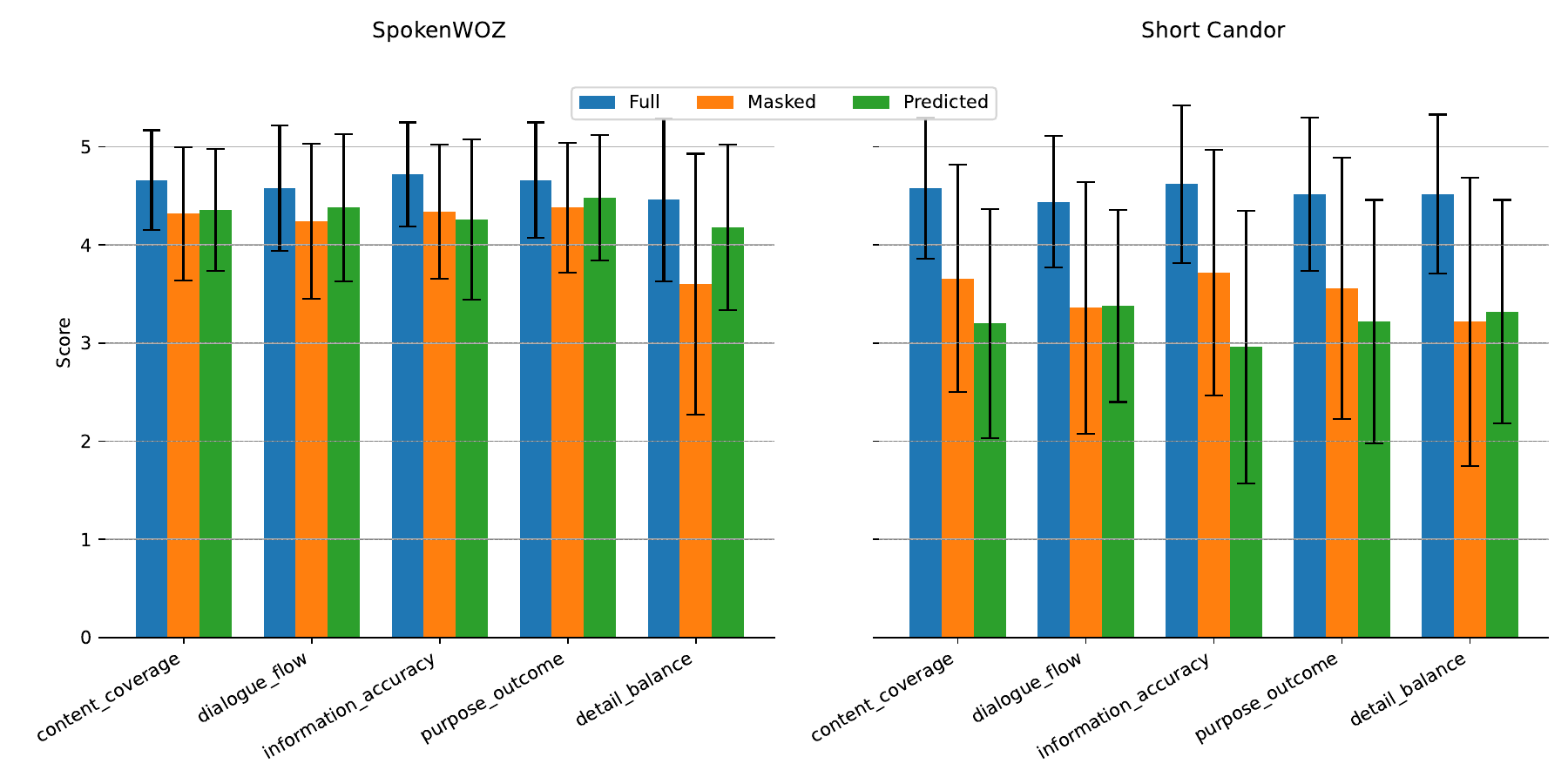}
  \caption{The results of our human rubric evaluation on SpokenWOZ and Candor.}
  \label{fig:humanspokensummrubric}
\end{figure*}

Fig.~\ref{fig:summrubric} shows the LLM rubric scores for our summarization task on text-based datasets.
Fig.~\ref{fig:humsummrubric} shows the human rubric scores for our summarization task on text-based datasets.
Fig.~\ref{fig:spokensummrubric} shows the LLM rubric scores for our summarization task on spoken-datasets.
Fig.~\ref{fig:humanspokensummrubric} shows the human rubric scores for our summarization task on spoken-datasets.

\section{Other Party Recreation Example} \label{apdx:oprex}
\textbf{Real Response}: okay, I have the Grafton Hotel Restaurant. What time would you like me to book your reservation?

\subsection{Base}
\textbf{Context:} Turn 1 [Speaker\_1]: I'm looking for a place to eat in the east that is expensive.
Turn 2 [Speaker\_2]: [MASKED]
Turn 3 [Speaker\_1]: How about an expensive British place? I'd like a reservation for Wednesday.
Turn 4 [Predict this turn : Speaker\_2]:

\textbf{Claude Response}: I found an expensive British restaurant in the east called XXXXXXX. What time would you like your reservation for Wednesday and how many people will be dining?

\textbf{Evaluation}: {'semantic\_similarity': 3, 'intent\_preservation': 5, 'precision': 0.0, 'recall': 0.0, 'anti\_hallucination\_score': 5, 'contextual\_appropriateness': 4, 'summary\_alignment': 3, 'actual\_specific\_info\_count': 1, 'xxx\_used\_count': 1, 'reasoning\_details': {'semantic\_similarity\_reasoning': 'Both suggest booking a British restaurant but differ in naming specifics.', 'semantic\_similarity': 3, 'intent\_preservation\_reasoning': 'Both responses aim to facilitate a reservation with similar inquiries.', 'intent\_preservation': 5, 'xxx\_masking\_compliance\_reasoning': 'Predicted response correctly uses XXXXXXX to mask the restaurant name.', 'xxx\_masking\_compliance': 5, 'contextual\_appropriateness\_reasoning': 'Predicted response continues reservation dialogue naturally.', 'contextual\_appropriateness': 4, 'summary\_alignment\_reasoning': 'Both summaries align in making a reservation but differ in details provided.', 'summary\_alignment': 3}, 'detail\_extraction': {'actual\_details': ['Grafton Hotel Restaurant'], 'predicted\_details': ['expensive British restaurant', 'XXXXXXX'], 'tp': 0, 'fp': 1, 'fn': 1, 'precision\_fraction': 0.0, 'recall\_fraction': 0.0}}

\subsection{Turn N+1}
\textbf{Context:} Turn 1 [Speaker\_1]: I'm looking for a place to eat in the east that is expensive.
Turn 2 [Speaker\_2]: [MASKED]
Turn 3 [Speaker\_1]: How about an expensive British place? I'd like a reservation for Wednesday.
Turn 4 [Predict this turn : Speaker\_2]: 
Turn 5 [Speaker\_1]: I would like a table at 14:00 on Wednesday for 5 people.

\textbf{Claude Response}: I found an expensive British restaurant in the east area. XXXXXXX serves British cuisine and is located at XXXXXXX. I can make a reservation for Wednesday. What time would you prefer and how many people will be dining?

\textbf{Evaluation}: {'semantic\_similarity': 2, 'intent\_preservation': 5, 'precision': 0.0, 'recall': 0.0, 'anti\_hallucination\_score': 4, 'contextual\_appropriateness': 4, 'summary\_alignment': 2, 'actual\_specific\_info\_count': 1, 'xxx\_used\_count': 2, 'reasoning\_details': {'semantic\_similarity\_reasoning': 'Responses mention booking a British restaurant but differ in specifics.', 'semantic\_similarity': 2, 'intent\_preservation\_reasoning': 'Both offer to assist in making a reservation.', 'intent\_preservation': 5, 'xxx\_masking\_compliance\_reasoning': 'Prediction masks specifics not in prior context.', 'xxx\_masking\_compliance': 4, 'contextual\_appropriateness\_reasoning': 'Predicted response smoothly continues the dialogue.', 'contextual\_appropriateness': 4, 'summary\_alignment\_reasoning': 'Summaries differ due to named locations and details.', 'summary\_alignment': 2}, 'detail\_extraction': {'actual\_details': ['Grafton Hotel Restaurant'], 'predicted\_details': ['British restaurant', 'XXXXXXX', 'British cuisine', 'XXXXXXX', 'Wednesday'], 'tp': 0, 'fp': 5, 'fn': 1, 'precision\_fraction': 0.0, 'recall\_fraction': 0.0}}

\subsection{Turn Lengths}
\textbf{Context:} Turn 1 [Speaker\_1]: I'm looking for a place to eat in the east that is expensive.
Turn 2 [Speaker\_2]: [MASKED - 20 words]
Turn 3 [Speaker\_1]: How about an expensive British place? I'd like a reservation for Wednesday.
Turn 4 [Predict this turn : Speaker\_2 - 17 Words]:

\textbf{Claude Response}: I found Grafton Hotel Restaurant, an expensive British restaurant in the east. What time and how many people for Wednesday?

\textbf{Evaluation}: {'semantic\_similarity': 4, 'intent\_preservation': 5, 'precision': 0.333, 'recall': 1, 'anti\_hallucination\_score': 2, 'contextual\_appropriateness': 3, 'summary\_alignment': 4, 'actual\_specific\_info\_count': 0, 'xxx\_used\_count': 0, 'reasoning\_details': {'semantic\_similarity\_reasoning': 'Both responses refer to Grafton Hotel Restaurant and ask about time.', 'semantic\_similarity': 4, 'intent\_preservation\_reasoning': 'Both responses aim to confirm details and request specifics.', 'intent\_preservation': 5, 'xxx\_masking\_compliance\_reasoning': 'Predicted response reveals details not explicit in context.', 'xxx\_masking\_compliance': 2, 'contextual\_appropriateness\_reasoning': 'Predicted response disrupts flow by adding unnecessary specifics.', 'contextual\_appropriateness': 3, 'summary\_alignment\_reasoning': 'Summaries generally align in seeking reservation details.', 'summary\_alignment': 4}, 'detail\_extraction': {'actual\_details': ['Grafton Hotel Restaurant', 'reservation', 'time'], 'predicted\_details': ['Grafton Hotel Restaurant', 'expensive British restaurant', 'east', 'Wednesday', 'time', 'people'], 'tp': 2, 'fp': 4, 'fn': 0, 'precision\_fraction': 0.333, 'recall\_fraction': 1}}

\subsection{Turn Lengths and Turn N+1}
\textbf{Context:} Turn 1 [Speaker\_1]: I'm looking for a place to eat in the east that is expensive.
Turn 2 [Speaker\_2]: [MASKED - 20 words]
Turn 3 [Speaker\_1]: How about an expensive British place? I'd like a reservation for Wednesday.
Turn 4 [Predict this turn : Speaker\_2 - 17 Words]: 
Turn 5 [Speaker\_1]: I would like a table at 14:00 on Wednesday for 5 people.

\textbf{Claude Response}: I found XXXXXXX, an expensive British restaurant in the east. What time and how many people?

\textbf{Evaluation}: {'semantic\_similarity': 3, 'intent\_preservation': 5, 'precision': 0.0, 'recall': 0.0, 'anti\_hallucination\_score': 5, 'contextual\_appropriateness': 5, 'summary\_alignment': 4, 'actual\_specific\_info\_count': 1, 'xxx\_used\_count': 1, 'reasoning\_details': {'semantic\_similarity\_reasoning': 'Responses suggest restaurant booking but differ in specificity.', 'semantic\_similarity': 3, 'intent\_preservation\_reasoning': 'Both responses aim to organize a reservation.', 'intent\_preservation': 5, 'xxx\_masking\_compliance\_reasoning': 'Prediction successfully masks specific restaurant name not in context.', 'xxx\_masking\_compliance': 5, 'contextual\_appropriateness\_reasoning': 'Predicted response appropriately follows inquiry about reservation.', 'contextual\_appropriateness': 5, 'summary\_alignment\_reasoning': 'Summaries aligned as both focus on reservation information.', 'summary\_alignment': 4}, 'detail\_extraction': {'actual\_details': ['Grafton Hotel Restaurant'], 'predicted\_details': ['XXXXXXX', 'an expensive British restaurant in the east'], 'tp': 0, 'fp': 2, 'fn': 1, 'precision\_fraction': 0.0, 'recall\_fraction': 0.0}}

\subsection{Local Context}
\textbf{Context:} Turn 3 [Speaker\_1]: How about an expensive British place? I'd like a reservation for Wednesday.
Turn 4 [Predict this turn : Speaker\_2]:
Turn 5 [Speaker\_1]: I would like a table at 14:00 on Wednesday for 5 people.

\textbf{Claude Response}: II found several expensive British restaurants available for Wednesday. XXXXXXX serves traditional British cuisine and has availability. What time would you prefer and how many people will be dining?

\textbf{Evaluation}: "semantic\_similarity": {"semantic\_similarity": 2,"intent\_preservation": 3,"precision": 0.0,"recall": 0.0,"anti\_hallucination\_score": 5,"contextual\_appropriateness": 4,"summary\_alignment": 2,"actual\_specific\_info\_count": 1, "xxx\_used\_count": 1 "reasoning\_details": {"semantic\_similarity\_reasoning": "Predicted focuses on options; actual specifies one place.", "semantic\_similarity": 2 "intent\_preservation\_reasoning": "Both offer help with reservations, differing in specificity.","intent\_preservation": 3, "xxx\_masking\_compliance\_reasoning": "Correct use of XXXXXXX for unknown restaurant name.","xxx\_masking\_compliance": 5,"contextual\_appropriateness\_reasoning": "Predicted asks for time and people; fits task flow.","contextual\_appropriateness": 4,"summary\_alignment\_reasoning": "Summaries diverge due to specificity difference.","summary\_alignment": 2},"detail\_extraction": {"actual\_details": ["Grafton Hotel Restaurant"],"predicted\_details": ["XXXXXXX serves traditional British cuisine"],"tp": 0,"fp": 1,"fn": 1,"precision\_fraction": 0.0,"recall\_fraction": 0.0}}

\section{Summarization Examples} \label{apdx:summex}

\subsection{Example 1: DailyDialog (Low Scores)}

\subsubsection{Full Conversation}
Turn 1 [Speaker\_1]: Are you all right ?
Turn 2 [Speaker\_2]: No , I'm not .
Turn 3 [Speaker\_1]: What's going on ?
Turn 4 [Speaker\_2]: I was just robbed !
Turn 5 [Speaker\_1]: Are you serious ?
Turn 6 [Speaker\_2]: Someone just robbed me .
Turn 7 [Speaker\_1]: Did he hurt you ?
Turn 8 [Speaker\_2]: He took all my things , but he didn't hurt me .
Turn 9 [Speaker\_1]: I can't believe that .
Turn 10 [Speaker\_2]: I need to make a statement at the police station .
Turn 11 [Speaker\_1]: I'll take you , if you'd like .
Turn 12 [Speaker\_2]: Thanks . 

\subsubsection{Oracle Summary}
This conversation involves Speaker\_1 checking on Speaker\_2's wellbeing after noticing something was wrong. Speaker\_2 reveals they have just been robbed, explaining that while the perpetrator took all their belongings, they were not physically harmed during the incident. Speaker\_1 expresses disbelief and concern about the robbery. The conversation concludes with Speaker\_2 stating they need to go to the police station to file a report, and Speaker\_1 offering to provide transportation there, which Speaker\_2 accepts with gratitude. The overall purpose is Speaker\_1 providing immediate emotional support and practical assistance to Speaker\_2 following a traumatic robbery incident.

\subsubsection{Reconstruction-Free Summary}
This conversation appears to be a concerned exchange where Speaker\_1 is checking on Speaker\_2's wellbeing after learning about some troubling situation. The dialogue begins with Speaker\_1 asking if Speaker\_2 is alright, then inquiring about what happened. Speaker\_1's responses suggest they are receiving increasingly serious information, as evidenced by their progression from asking "What's going on?" to expressing disbelief with "Are you serious?" and "I can't believe that." A key concern emerges when Speaker\_1 specifically asks "Did he hurt you?" indicating that Speaker\_2 may have experienced some form of harm or mistreatment from another person. The conversation concludes with Speaker\_1 offering practical support by saying "I'll take you, if you'd like," suggesting they are willing to transport Speaker\_2 somewhere, possibly for help, safety, or to address whatever situation has occurred. Overall, this appears to be a supportive conversation where one person is offering assistance to someone who may have experienced a difficult or potentially harmful situation.

\subsubsection{Reconstruction-Heavy Summary}
This conversation involves Speaker\_1 checking on Speaker\_2's wellbeing after noticing they don't seem alright. Speaker\_2 reveals they are not feeling well and discloses that they are experiencing XXXXXXX, which appears to involve a physical altercation where someone hit them that morning. While Speaker\_2 indicates they weren't seriously hurt, they explain that the person also made threats and said they would return. Speaker\_1 expresses disbelief and concern about the situation. Speaker\_2 decides they need medical attention and wants to go to the hospital to get checked out, and Speaker\_1 offers to provide transportation, which Speaker\_2 gratefully accepts. The conversation centers around Speaker\_2 seeking help and support after what appears to be a threatening encounter, with Speaker\_1 providing immediate assistance and care.

\subsubsection{Rubric Scores}
\textbf{Oracle:} content\_coverage\_reasoning: Accurately covers robbery and police station visit., content\_coverage: 5, dialogue\_flow\_reasoning: Reflects natural progression and interaction accurately., dialogue\_flow: 5, information\_accuracy\_reasoning: Faithfully represents the dialogue's content., information\_accuracy: 5, purpose\_outcome\_reasoning: Clearly conveys the robbery and police report outcome., purpose\_outcome: 5, detail\_balance\_reasoning: Balances details from both speakers effectively., detail\_balance: 5

\textbf{Reconstruction-Free:} content\_coverage\_reasoning: Misses key details like robbery and police station visit., content\_coverage: 2, dialogue\_flow\_reasoning: Captures progression of concern but lacks specific robbery details., dialogue\_flow: 3, information\_accuracy\_reasoning: Inaccurate as it omits robbery and police station details., information\_accuracy: 2, purpose\_outcome\_reasoning: Fails to clearly convey the robbery and police report outcome., purpose\_outcome: 2, detail\_balance\_reasoning: Focuses more on Speaker\_1's concern, less on Speaker\_2's situation., detail\_balance: 2

\textbf{Reconstruction-Heavy}: content\_coverage\_reasoning: Introduces incorrect elements like physical altercation and hospital visit., content\_coverage: 1, dialogue\_flow\_reasoning: Flow is disrupted by inaccurate events and outcomes., dialogue\_flow: 1, information\_accuracy\_reasoning: Contains significant inaccuracies about the dialogue content., information\_accuracy: 1, purpose\_outcome\_reasoning: Misrepresents the dialogue's purpose and outcome., purpose\_outcome: 1, detail\_balance\_reasoning: Incorrect focus on non-existent events, unbalanced details., detail\_balance: 1

\subsubsection{Ranking}
The ranking is:
\begin{enumerate}
    \item Oracle
    \item Reconstruction-free
    \item Reconstruction-heavy
\end{enumerate}

\subsection{Example 2: MultiWOZ (Average Scores)}

\subsubsection{Full Conversation}
Turn 1 [Speaker\_1]: Can you tell me about any hungarian restaurants in the centre?
Turn 2 [Speaker\_2]: I'm sorry I do not have any Hungarian restaurants in Cambridge. Is there another type of cuisine you might be interested in?
Turn 3 [Speaker\_1]: How about one that serves modern european food?
Turn 4 [Speaker\_2]: We have several in the center, and one in the south. They are in assorted price ranges. Do you have any preferences?
Turn 5 [Speaker\_1]: Whatever you recommend, please book me for 5 people at 12:15 on friday.
Turn 6 [Speaker\_2]: I have made a booking for De Luca Cucina and bar. The table will be reserved for 15 minutes. Reference number is : 2X1IGK9D .
Turn 7 [Speaker\_1]: I also need to find a two star room.
Turn 8 [Speaker\_2]: How about the ashley hotel? They are moderately priced and have both internet and parking.
Turn 9 [Speaker\_1]: That sounds great. Could you give me the address please?
Turn 10 [Speaker\_2]: Sure, the address is 74 chesterton road post code cb41er. Would you like a reservation?
Turn 11 [Speaker\_1]: No thanks. I do want a taxi though to go from the hotel to the restaurant, arriving by my reservation time. Tell me the car type and contact number.
Turn 12 [Speaker\_2]: I've booked a grey bmw, phone number 07260290005. Will that be all?
Turn 13 [Speaker\_1]: That is everything, thanks for you help.
Turn 14 [Speaker\_2]: Great. Enjoy your stay!

\subsubsection{Oracle Summary}
The conversation involves a customer seeking dining and accommodation assistance in Cambridge. The customer initially inquired about Hungarian restaurants in the centre, but when informed that none were available, they switched to requesting modern European cuisine. The assistant offered several options and the customer asked for a recommendation, requesting a booking for 5 people at 12:15 on Friday. The assistant successfully booked a table at De Luca Cucina and bar with reference number 2X1IGK9D, noting the table would be reserved for 15 minutes. The customer then requested a two-star hotel room, and the assistant recommended the Ashley Hotel, describing it as moderately priced with internet and parking amenities. After the customer requested the address (74 Chesterton Road, postcode CB41ER), they declined making a hotel reservation but asked for taxi transportation from the hotel to the restaurant, arriving by their reservation time. The assistant arranged a grey BMW taxi with contact number 07260290005, completing all the customer's requests for their dining, accommodation, and transportation needs in Cambridge.

\subsubsection{Reconstruction-Free Summary}
This conversation involves a customer seeking assistance with restaurant reservations, hotel accommodations, and transportation arrangements in what appears to be a city center area. The customer initially inquires about Hungarian restaurants in the centre, then shifts to asking about modern European cuisine options. They proceed to make a restaurant reservation for 5 people at 12:15 on Friday, accepting whatever recommendation the assistant provides. The conversation then expands to include accommodation needs, with the customer requesting a two-star hotel room and asking for the address details. Finally, the customer arranges taxi transportation from the hotel to the restaurant, ensuring they arrive by their reservation time, and requests information about the car type and contact number for the taxi service. The interaction concludes with the customer expressing satisfaction that all their travel and dining arrangements have been handled.

\subsubsection{Reconstruction-Heavy Summary}
This conversation involves a customer seeking restaurant and accommodation recommendations in the centre area, along with transportation arrangements. The customer initially inquires about Hungarian restaurants but then shifts their preference to modern European cuisine. The assistant provides restaurant options and successfully books a table for 5 people at 12:15 on Friday at a recommended modern European restaurant, providing a reference number for the reservation. The customer then requests a two-star hotel room, and the assistant finds a suitable two-star hotel in the centre area and provides the address when requested. Finally, the customer arranges taxi transportation from the hotel to the restaurant, timed to arrive by their reservation time, and receives details about the car type and contact number. The conversation concludes with all the customer's needs being met - dining reservation, accommodation, and transportation - creating a complete travel arrangement for their Friday plans.

\subsubsection{Rubric Scores}
\textbf{Oracle:} content\_coverage\_reasoning: Includes specific names and details, covering all key points., content\_coverage: 5, dialogue\_flow\_reasoning: Reflects the natural progression of the dialogue well., dialogue\_flow: 5, information\_accuracy\_reasoning: Highly accurate with specific details., information\_accuracy: 5, purpose\_outcome\_reasoning: Conveys the dialogue's goals and results clearly., purpose\_outcome: 5, detail\_balance\_reasoning: Balances details from both speakers effectively., detail\_balance: 5

\textbf{Reconstruction-Free:} content\_coverage\_reasoning: Covers main actions but lacks specific details like names., content\_coverage: 3, dialogue\_flow\_reasoning: Captures the sequence but lacks some interaction nuances., dialogue\_flow: 3, information\_accuracy\_reasoning: Accurate but omits specific details., information\_accuracy: 3, purpose\_outcome\_reasoning: Conveys the overall purpose but lacks detail., purpose\_outcome: 4, detail\_balance\_reasoning: Balances speakers' contributions but lacks detail., detail\_balance: 3

\textbf{Reconstruction-Heavy}: content\_coverage\_reasoning: Misses specific restaurant and hotel names, but covers all main actions., content\_coverage: 4, dialogue\_flow\_reasoning: Captures the sequence of requests and responses well., dialogue\_flow: 4, information\_accuracy\_reasoning: Accurate but lacks specific names and details., information\_accuracy: 3, purpose\_outcome\_reasoning: Clearly conveys the goals and outcomes of the dialogue., purpose\_outcome: 5, detail\_balance\_reasoning: Balances both speakers' contributions adequately., detail\_balance: 4

\subsection{Example 3: Candor (Average Scores)}

\subsubsection{Full Conversation}
Turn 1 [Speaker\_1]: Yeah. Mhm, Mhm. Mhm, mm. Right.
Turn 2 [Speaker\_2]: steve
Turn 3 [Speaker\_1]: Hello?
Turn 4 [Speaker\_2]: hi
Turn 5 [Speaker\_1]: Mm hmm. Nice to meet you.
Turn 6 [Speaker\_2]: meet you. I'm
Turn 7 [Speaker\_1]: I'm a, I'm from new york.
Turn 8 [Speaker\_2]: from new york,
Turn 9 [Speaker\_1]: Mhm.
Turn 10 [Speaker\_2]: in California but I'm from Oregon
Turn 11 [Speaker\_1]: Okay. Um yeah, it's like 11 PM right now in new york, so yeah.
Turn 12 [Speaker\_2]: um
Turn 13 [Speaker\_1]: Mhm
Turn 14 [Speaker\_2]: family on that side of the country and I forget about the time difference.
Turn 15 [Speaker\_1]: uh yeah, so it's eight PM in Cali oh, okay.
Turn 16 [Speaker\_2]: It's still like still decently early.
Turn 17 [Speaker\_1]: Yeah, yeah, mm.
Turn 18 [Speaker\_2]: I just
Turn 19 [Speaker\_1]: Mhm.
Turn 20 [Speaker\_2]: a question in my
Turn 21 [Speaker\_1]: Right.
Turn 22 [Speaker\_2]: completely
Turn 23 [Speaker\_1]: Mhm. Mhm.
Turn 24 [Speaker\_2]: Have you done this survey before?
Turn 25 [Speaker\_1]: No, it's my first time.
Turn 26 [Speaker\_2]: Is your first time?
Turn 27 [Speaker\_1]: Yeah. How about, you know?
Turn 28 [Speaker\_2]: third time
Turn 29 [Speaker\_1]: Oh wow.
Turn 30 [Speaker\_2]: like the third time
Turn 31 [Speaker\_1]: Mhm. Oh wow, that's you could, that's a lot.
Turn 32 [Speaker\_2]: yeah I think so. This is my first round doing it. I think you can do it up to like six times
Turn 33 [Speaker\_1]: Mhm.
Turn 34 [Speaker\_2]: works because
Turn 35 [Speaker\_1]: Mhm.
Turn 36 [Speaker\_2]: out of town this week so like
Turn 37 [Speaker\_1]: Uh huh.
Turn 38 [Speaker\_2]: to do
Turn 39 [Speaker\_1]: Okay, so it, Right, I'm just wait, so it's six times you could do it in the this round.
Turn 40 [Speaker\_2]: Um so I messaged like what is it like you know how they
Turn 41 [Speaker\_1]: Oh,
Turn 42 [Speaker\_2]: you to schedule to schedule
Turn 43 [Speaker\_1]: mm hmm.
Turn 44 [Speaker\_2]: I messaged it with a question
Turn 45 [Speaker\_1]: Mhm.
Turn 46 [Speaker\_2]: that I think it's six total
Turn 47 [Speaker\_1]: I'm interesting. Okay. Yeah,
Turn 48 [Speaker\_2]: kind of weird
Turn 49 [Speaker\_1]: no,
Turn 50 [Speaker\_2]: worded it.
Turn 51 [Speaker\_1]: yeah,
Turn 52 [Speaker\_2]: like six total in a lifetime and be awesome
Turn 53 [Speaker\_1]: cool.
Turn 54 [Speaker\_2]: around though.
Turn 55 [Speaker\_1]: Mhm. Um how is, how is county now with like, I guess the air and stuff
Turn 56 [Speaker\_2]: um The air is good
Turn 57 [Speaker\_1]: overwhelmed?
Turn 58 [Speaker\_2]: of fires
Turn 59 [Speaker\_1]: Yeah
nTurn 60 [Speaker\_2]: that we're all in the
Turn 61 [Speaker\_1]: I guess it depends on like how close you are to L. A. Well. Mhm.
Turn 62 [Speaker\_2]: hear it. I know I'm about 35 miles from san
Turn 63 [Speaker\_1]: Yeah.
Turn 64 [Speaker\_2]: I'm not sure the exact distance on L. A. But I know when the fires going on were
Turn 65 [Speaker\_1]: One.
Turn 66 [Speaker\_2]: away and it was scary.
Turn 67 [Speaker\_1]: Yeah. Okay. That is kind of close still,
Turn 68 [Speaker\_2]: some of the stories I read were so heartbreaking
Turn 69 [Speaker\_1]: wow.
Turn 70 [Speaker\_2]: I'm coming on out of nowhere. It scared
Turn 71 [Speaker\_1]: Mhm. Right. My God
Turn 72 [Speaker\_2]: for the people who lost their homes like
Turn 73 [Speaker\_1]: many, yeah.
Turn 74 [Speaker\_2]: and imagine that at all.
Turn 75 [Speaker\_1]: Hope you have like good insurance. Right. Yeah.
Turn 76 [Speaker\_2]: lucky we live on a military base
Turn 77 [Speaker\_1]: Yeah.
Turn 78 [Speaker\_2]: we can just kind of go on to the next,
Turn 79 [Speaker\_1]: Oh, come on,
Turn 80 [Speaker\_2]: those like I don't know how you rebuild your life after
Turn 81 [Speaker\_1]: wow, that is this crazy? Mhm.
Turn 82 [Speaker\_2]: so what do you do, do you go to school or do you
Turn 83 [Speaker\_1]: Yeah, I'm working at schools. I'm a therapy. Okay,
Turn 84 [Speaker\_2]: Oh what kind of
Turn 85 [Speaker\_1]: occupational therapist.
Turn 86 [Speaker\_2]: you're a lifesaver? My daughter
Turn 87 [Speaker\_1]: Yeah.
Turn 88 [Speaker\_2]: goes to ot
Turn 89 [Speaker\_1]: Unharmed.
Turn 90 [Speaker\_2]: a
Turn 91 [Speaker\_1]: Oh, in schools?
Turn 92 [Speaker\_2]: Um So not
Turn 93 [Speaker\_1]: Mhm,
Turn 94 [Speaker\_2]: in
Turn 95 [Speaker\_1]: mm hmm.
Turn 96 [Speaker\_2]: qualify for like in part um in
Turn 97 [Speaker\_1]: Mhm.
Turn 98 [Speaker\_2]: R.
Turn 99 [Speaker\_1]: Mhm.
Turn 100 [Speaker\_2]: But she does private ot
Turn 101 [Speaker\_1]: Okay, wow.
Turn 102 [Speaker\_2]: for the Ot
Turn 103 [Speaker\_1]: Mhm.
Turn 104 [Speaker\_2]: speech
Turn 105 [Speaker\_1]: Yeah.
Turn 106 [Speaker\_2]: and physical
Turn 107 [Speaker\_1]: Okay. Yeah. Those students who have all those are super busy.
Turn 108 [Speaker\_2]: Yeah. Yeah, pretty
Turn 109 [Speaker\_1]: Yeah. Like after after school, Right. So that means that your, you have to take her to those places.
Turn 110 [Speaker\_2]: Yeah, so with Covid she's just starting school, she just started preschool.
Turn 111 [Speaker\_1]: Mhm.
Turn 112 [Speaker\_2]: yeah usually it's after school, but luckily school right now it's like 30 minutes to maybe 60 minutes a day, so
Turn 113 [Speaker\_1]: Yeah.
Turn 114 [Speaker\_2]: too bad
Turn 115 [Speaker\_1]: Right,
Turn 116 [Speaker\_2]: it definitely keeps her busy a couple days a
Turn 117 [Speaker\_1]: wow. Okay. I was gonna say like you look really young and I was like, you have a daughter?
Turn 118 [Speaker\_2]: young.
Turn 119 [Speaker\_1]: Okay.
Turn 120 [Speaker\_2]: I had her, I found out I was pregnant with her right after I turned 19.
Turn 121 [Speaker\_1]: Okay. Wait, that means I'm ordered, I'm 26.
Turn 122 [Speaker\_2]: I'm 24 I'll be 25 in me.
Turn 123 [Speaker\_1]: Oh, okay. Yeah. Okay. I think. Okay. You are younger than me.
Turn 124 [Speaker\_2]: where I live,
Turn 125 [Speaker\_1]: Mhm.
Turn 126 [Speaker\_2]: lot of young moms, but like when we're not on base, a lot of people like how old do you like especially
Turn 127 [Speaker\_1]: Yeah.
Turn 128 [Speaker\_2]: when I found that if she's not with me, they're like what?
Turn 129 [Speaker\_1]: Right. Yeah, I mean,
Turn 130 [Speaker\_2]: don't blame you at all.
Turn 131 [Speaker\_1]: Mhm.
Turn 132 [Speaker\_2]: I had her super young,
Turn 133 [Speaker\_1]: Oh,
Turn 134 [Speaker\_2]: change it now
Turn 135 [Speaker\_1]: mm hmm. Are you, I guess planning for another kid?
Turn 136 [Speaker\_2]: Um Kind of yes, eventually um we originally wanted a kid a little bit closer, so she'll be five in december.
Turn 137 [Speaker\_1]: Uh huh.
Turn 138 [Speaker\_2]: um I had a really really hard pregnancy with her and a really hard delivery
Turn 139 [Speaker\_1]: Oh
nTurn 140 [Speaker\_2]: unfortunately something about my body after having her, I've had three surgeries due
Turn 141 [Speaker\_1]: mm.
Turn 142 [Speaker\_2]: that keep coming up
Turn 143 [Speaker\_1]: Oh,
Turn 144 [Speaker\_2]: and it's just it's a
Turn 145 [Speaker\_1]: Oh my gosh, Yeah.
Turn 146 [Speaker\_2]: I have the weirdest like when she was 18 months, I had a mass the size of a lime
Turn 147 [Speaker\_1]: Doing that's so, yeah, that's big.
Turn 148 [Speaker\_2]: Yeah, on my C. Section scar,
Turn 149 [Speaker\_1]: Oh,
Turn 150 [Speaker\_2]: got rid of it.
Turn 151 [Speaker\_1]: okay.
Turn 152 [Speaker\_2]: then last december I had the same kind of pain and scar tissue on my C section scar had literally like blinded
Turn 153 [Speaker\_1]: Oh,
Turn 154 [Speaker\_2]: like organs together, and then I'm getting ready to go through the same thing
Turn 155 [Speaker\_1]: blanca. Yeah.
Turn 156 [Speaker\_2]: want another, but
Turn 157 [Speaker\_1]: Right,
Turn 158 [Speaker\_2]: maybe a little bit.
Turn 159 [Speaker\_1]: okay. Yeah. Actually you're still pretty young, so yeah.
Turn 160 [Speaker\_2]: too. Like when we first, when we first had her were really tight on money and just the older that I get, I think the better I do as a mom,
Turn 161 [Speaker\_1]: Mhm.
Turn 162 [Speaker\_2]: not that I a bad mom when I first had her, but looking back I didn't know just how inexperienced I really was
Turn 163 [Speaker\_1]: Nine.
Turn 164 [Speaker\_2]: and so many things were harder on both of us, I think.
Turn 165 [Speaker\_1]: Mhm.
Turn 166 [Speaker\_2]: I think with a newborn I'll do better like anywhere from like now in the next few years versus when I had her,
Turn 167 [Speaker\_1]: Yeah the second time.
Turn 168 [Speaker\_2]: think you're young,
Turn 169 [Speaker\_1]: Huh?
Turn 170 [Speaker\_2]: you don't think like your unexperienced until you look back a couple years
Turn 171 [Speaker\_1]: Why?
Turn 172 [Speaker\_2]: oh
Turn 173 [Speaker\_1]: Exactly? But yes I think you should always be better. Okay well
Turn 174 [Speaker\_2]: kids eventually?
Turn 175 [Speaker\_1]: yeah I mean a few years to Mhm.
Turn 176 [Speaker\_2]: are no a shame in waiting because it's so much harder to like if you have any career goals or personal goals to get those going, if you don't already have them started, which
Turn 177 [Speaker\_1]: Mhm
Turn 178 [Speaker\_2]: when I have my daughter,
Turn 179 [Speaker\_1]: Uh huh
Turn 180 [Speaker\_2]: so much harder to start all that stuff.
Turn 181 [Speaker\_1]: Oh what do we do now?
Turn 182 [Speaker\_2]: Um So I stay home with her right now mostly because so she has
Turn 183 [Speaker\_1]: Oh.
Turn 184 [Speaker\_2]: therapy now then she used to, she was born really, really small
Turn 185 [Speaker\_1]: Mhm.
Turn 186 [Speaker\_2]: so she keeps me busy and um the first three years of her life we lived in Japan and there weren't
Turn 187 [Speaker\_1]: Oh
Turn 188 [Speaker\_2]: lot of jobs at all.
Turn 189 [Speaker\_1]: wow
Turn 190 [Speaker\_2]: Yeah.
Turn 191 [Speaker\_1]: that's cool, wow.
Turn 192 [Speaker\_2]: and I said I want to go
Turn 193 [Speaker\_1]: Which part of japan?
Turn 194 [Speaker\_2]: Okinawa.
Turn 195 [Speaker\_1]: Oh wow I only been to Tokyo.
Turn 196 [Speaker\_2]: see
Turn 197 [Speaker\_1]: Oh
Turn 198 [Speaker\_2]: you know I never made it to Tokyo my husband did, I was really jealous,
Turn 199 [Speaker\_1]: so
Turn 200 [Speaker\_2]: it was crazy
Turn 201 [Speaker\_1]: yeah it's awesome, so three years in japan adjourns husband station there. Right wow
Turn 202 [Speaker\_2]: Yes,
Turn 203 [Speaker\_1]: wow. How can my, what is that like mountain is there? What is that?
Turn 204 [Speaker\_2]: no it's an island,
Turn 205 [Speaker\_1]: Uh huh.
Turn 206 [Speaker\_2]: it's not very mountainous at all. Very Americanized.
Turn 207 [Speaker\_1]: Oh okay
Turn 208 [Speaker\_2]: almost like I describe it like they had like kind of this cute culture of like american things, they're like they had stores that sold like american products and we're like, it felt like you were in the sixties, a lot of like the island
Turn 209 [Speaker\_1]: wow I think
Turn 210 [Speaker\_2]: would have like american diners and
Turn 211 [Speaker\_1]: oh
Turn 212 [Speaker\_2]: but it was, it's weird when that's what you like are used to and then you see like their version.
Turn 213 [Speaker\_1]: yeah
Turn 214 [Speaker\_2]: Yeah.
Turn 215 [Speaker\_1]: there are a lot of japanese people living there or Okay that's cool, I'm gonna search
Turn 216 [Speaker\_2]: on base but it was really nice. They're really
Turn 217 [Speaker\_1]: oh what did you, where did you go after japan?
Turn 218 [Speaker\_2]: uh here to
Turn 219 [Speaker\_1]: No okay. For the next 10 years. Right.
Turn 220 [Speaker\_2]: I
Turn 221 [Speaker\_1]: Mhm.
Turn 222 [Speaker\_2]: want to go back to Japan. My husband's not so sure. So they always want people to go there because a lot of people, so a lot of like the single guys and women
Turn 223 [Speaker\_1]: Yeah.
Turn 224 [Speaker\_2]: who are serving want to go there. But it seems like once you have a family, most people don't because
Turn 225 [Speaker\_1]: Right
Turn 226 [Speaker\_2]: from everyone,
Turn 227 [Speaker\_1]: yeah
Turn 228 [Speaker\_2]: but I'd rather be there with a kid, it's so much safer.
Turn 229 [Speaker\_1]: yeah that's true
Turn 230 [Speaker\_2]: Yeah and the japanese locals are so good with kid
nTurn 231 [Speaker\_1]: yeah
Turn 232 [Speaker\_2]: and so precious. Just the whole culture there is so different
Turn 233 [Speaker\_1]: yeah.
Turn 234 [Speaker\_2]: in America.
Turn 235 [Speaker\_1]: Did you learn japanese? No.
Turn 236 [Speaker\_2]: wish I would have tried more
Turn 237 [Speaker\_1]: Oh. Mhm.
Turn 238 [Speaker\_2]: you didn't really like looking back. I wish I would have known just to communicate better especially with just how great everyone was there. But it wasn't needed as much as other parts of Japan I think because it was so like so much of an american presence.
Turn 239 [Speaker\_1]: Oh wow okay.
Turn 240 [Speaker\_2]: Yeah
Turn 241 [Speaker\_1]: Oh.
Turn 242 [Speaker\_2]: every local we met new at least a little like english.
Turn 243 [Speaker\_1]: Oh wow. Oh interesting. All right.
Turn 244 [Speaker\_2]: but I've heard the areas around it not as much like definitely need to have a translation app more
Turn 245 [Speaker\_1]: Yeah, I don't,
Turn 246 [Speaker\_2]: I would have learned a little
Turn 247 [Speaker\_1]: did you know, did you learn like any phrases like high,
Turn 248 [Speaker\_2]: that I can't remember remember
Turn 249 [Speaker\_1]: mm
Turn 250 [Speaker\_2]: we're like a couple of small ones but
Turn 251 [Speaker\_1]: konichiwa? Hello? Yes.
Turn 252 [Speaker\_2]: it hasn't been that long, it just feels like since my daughter was born there in the first like two years of her life just went by.
Turn 253 [Speaker\_1]: Yeah.
Turn 254 [Speaker\_2]: I wish I would have immersed myself more in the culture there because at
Turn 255 [Speaker\_1]: Oh.
Turn 256 [Speaker\_2]: I wasn't that happy about being there.
Turn 257 [Speaker\_1]: Oh yeah, because you're
Turn 258 [Speaker\_2]: your whole life.
Turn 259 [Speaker\_1]: yes, yeah, queens.
Turn 260 [Speaker\_2]: queen? Oh
Turn 261 [Speaker\_1]: Mhm.
Turn 262 [Speaker\_2]: what,
Turn 263 [Speaker\_1]: Yeah and it's the same, it's like living in, okay, so I visited Cali like L. A. And san Fran area and I think it would be like more akin to san Fran area because queens is because it's like less carbonated and more housing and stuff. Um I love it. Okay.
Turn 264 [Speaker\_2]: want to visit new york. I've only driven through
Turn 265 [Speaker\_1]: Oh where did you go
Turn 266 [Speaker\_2]: Um I drove through, I don't know where I drove there, it was on my way to new is from north Carolina to new Hampshire
Turn 267 [Speaker\_1]: life?
Turn 268 [Speaker\_2]: it was just, it was so quick,
Turn 269 [Speaker\_1]: Um
Turn 270 [Speaker\_2]: song
Turn 271 [Speaker\_1]: Oh
Turn 272 [Speaker\_2]: the next time we go that way I want to stop
Turn 273 [Speaker\_1]: yeah, for sure, it's great. Where have you travelled to?
Turn 274 [Speaker\_2]: Um No I thought many places honestly I've lived so I lived in Japan for three years. Um I grew up in Oregon in Portland until I was 16
Turn 275 [Speaker\_1]: Mhm.
Turn 276 [Speaker\_2]: and then I moved to south Carolina for a year
Turn 277 [Speaker\_1]: Mhm
Turn 278 [Speaker\_2]: north Carolina and then japan,
Turn 279 [Speaker\_1]: Oh
Turn 280 [Speaker\_2]: that many
Turn 281 [Speaker\_1]: okay. But that be, that's cool that you're in japan, I mean you went to japan? Mhm
Turn 282 [Speaker\_2]: that's it. What
Turn 283 [Speaker\_1]: mhm
Turn 284 [Speaker\_2]: you? Where all have you traveled?
Turn 285 [Speaker\_1]: uh
Turn 286 [Speaker\_2]: a little
Turn 287 [Speaker\_1]: yeah, I travel so many places but it is like, like to travel but um like new york, I mean went to Barcelona and Portugal and London I enroll uh huh.
Turn 288 [Speaker\_2]: are the Ireland like I want to go?
Turn 289 [Speaker\_1]: God, it's a, it's like very small, small, let's about small town, like everything in like a day and a half, it's really tiny but it's so nice, it's very nice. Um then I went to the asian countries like china Korea Hong kong is sarah okay.
Turn 290 [Speaker\_2]: so awesome.
Turn 291 [Speaker\_1]: Yeah.
Turn 292 [Speaker\_2]: I feel like when I was like in high school, so I didn't, I had a really weird childhood that was just very very very sheltered and you think when I turned 18, like I want to go crazy exploring but the idea was so foreign to me,
Turn 293 [Speaker\_1]: Mhm
Turn 294 [Speaker\_2]: I honestly like almost kind of shunned it and the older I get, it's like it's not an option to travel a whole bunch right now,
Turn 295 [Speaker\_1]: mhm
Turn 296 [Speaker\_2]: think so much more of an urge to travel
Turn 297 [Speaker\_1]: We do, wow.
Turn 298 [Speaker\_2]: if it's just like little places within
Turn 299 [Speaker\_1]: Yeah. Right. Right. You can go to Maine and no.
Turn 300 [Speaker\_2]: I want to go to me and we have family near there.
Turn 301 [Speaker\_1]: Yeah.
Turn 302 [Speaker\_2]: laws. Um, my mother in law lives in new Hampshire, but they visit me in a lot and have family there. And then my father in law side of the family lives in Colorado
Turn 303 [Speaker\_1]: Oh
Turn 304 [Speaker\_2]: It's really pretty there.
Turn 305 [Speaker\_1]: yeah, yeah.
Turn 306 [Speaker\_2]: have a cabin there, that's really
Turn 307 [Speaker\_1]: Yeah. Do you like to go there in the winter word?
Turn 308 [Speaker\_2]: haven't been in the winter yet. We've only been in summer.
Turn 309 [Speaker\_1]: Okay.
Turn 310 [Speaker\_2]: hoping to try and go this year. But with Covid traveling kind of
Turn 311 [Speaker\_1]: Right. Yeah.
Turn 312 [Speaker\_2]: It's hard. And with my husband's job that he's not allowed to travel more than like 400 miles away right now.
Turn 313 [Speaker\_1]: Oh,
Turn 314 [Speaker\_2]: So that kind of until Covid's
Turn 315 [Speaker\_1]: what?
Turn 316 [Speaker\_2]: military seems to be on travel restrictions at least here,
Turn 317 [Speaker\_1]: Right, okay.
Turn 318 [Speaker\_2]: Which is good.
Turn 319 [Speaker\_1]: Yeah.
Turn 320 [Speaker\_2]: like you don't, you don't
Turn 321 [Speaker\_1]: Yeah.
Turn 322 [Speaker\_2]: want to do stuff until you're told you can't do it
Turn 323 [Speaker\_1]: Right, right.
Turn 324 [Speaker\_2]: at least not as much.
Turn 325 [Speaker\_1]: Mhm
Turn 326 [Speaker\_2]: is new york
Turn 327 [Speaker\_1]: Right.
Turn 328 [Speaker\_2]: prove ID?
Turn 329 [Speaker\_1]: Um it was like the numbers are really high here. Really? Really what?
Turn 330 [Speaker\_2]: coronavirus first hit new york and kelly kept like fighting each other for the top
Turn 331 [Speaker\_1]: Yeah. Yeah and
Turn 332 [Speaker\_2]: many cases.
Turn 333 [Speaker\_1]: yeah, but it might not have been okay. I mean like I've been going out, I mean of course everyone has a mask on, everyone's pretty good. Like some people like cases are rising but not many people are dying by guys.
Turn 334 [Speaker\_2]: noticed that there's less
Turn 335 [Speaker\_1]: Oh,
Turn 336 [Speaker\_2]: worries about the more extreme sides now.
Turn 337 [Speaker\_1]: mm.
Turn 338 [Speaker\_2]: though like for example, here, the numbers have been rising pretty good too.
Turn 339 [Speaker\_1]: Mhm.
Turn 340 [Speaker\_2]: there's not as many really hard stories. Like, you know, it still happens,
Turn 341 [Speaker\_1]: Three
Turn 342 [Speaker\_2]: I feel like you heard more at the beginning.
Turn 343 [Speaker\_1]: Yeah. Crazy. In the beginning it was like so many people dying. No. Yes. I don't know. It's not that many.
Turn 344 [Speaker\_2]: I don't know what to think
Turn 345 [Speaker\_1]: Yeah.
Turn 346 [Speaker\_2]: all of it. I'm not super
Turn 347 [Speaker\_1]: Yeah.
Turn 348 [Speaker\_2]: educated on it. But I also know there's so many opinions on
Turn 349 [Speaker\_1]: Yeah. Mhm.
Turn 350 [Speaker\_2]: So it's hard to know like I definitely believe it's real. I don't
Turn 351 [Speaker\_1]: Yeah.
Turn 352 [Speaker\_2]: people who are like completely
Turn 353 [Speaker\_1]: Uh huh.
Turn 354 [Speaker\_2]: a couple
Turn 355 [Speaker\_1]: No.
Turn 356 [Speaker\_2]: and neighbors who are like, it's a scam. Don't wear your mask, like the
Turn 357 [Speaker\_1]: What?
Turn 358 [Speaker\_2]: you. And I'm like,
Turn 359 [Speaker\_1]: Uh huh.
Turn 360 [Speaker\_2]: it's that far. Like,
Turn 361 [Speaker\_1]: Mhm.
Turn 362 [Speaker\_2]: this.
Turn 363 [Speaker\_1]: Yeah. And the mask is like, you know, with the mass, it's not like they're putting a tracking device on you, you know?
Turn 364 [Speaker\_2]: It's not some crazy some people treat it like it is like for
Turn 365 [Speaker\_1]: Yeah.
Turn 366 [Speaker\_2]: um, I work very, very, very part time a church we go to and a couple of months ago. Um, so they had the assistant,
Turn 367 [Speaker\_1]: Yeah.
Turn 368 [Speaker\_2]: Ministry
Turn 369 [Speaker\_1]: Yeah.
Turn 370 [Speaker\_2]: where I work
Turn 371 [Speaker\_1]: Mhm.
Turn 372 [Speaker\_2]: and I ended up giving my notice because they were literally like not
Turn 373 [Speaker\_1]: Oh
Turn 374 [Speaker\_2]: abiding by any like
Turn 375 [Speaker\_1]: hi,
Turn 376 [Speaker\_2]: letting us have Lysol wipes in the room,
Turn 377 [Speaker\_1]: mm.
Turn 378 [Speaker\_2]: like letting us wear masks, like literally like making fun of you if you tried
Turn 379 [Speaker\_1]: Oh my gosh, that's your church.
Turn 380 [Speaker\_2]: I left, I gave a notice because they didn't have, they were having child care and like services when they weren't supposed to. So I gave notice after coming back and then they switched like those pastors left and now the pastors that are in charge of taking it seriously. So I work like two days a week
Turn 381 [Speaker\_1]: Oh, okay.
Turn 382 [Speaker\_2]: regretfully precautions but like the way they made fun of it before I was like, I don't know if I want to be in this
Turn 383 [Speaker\_1]: Yeah, that's that nice. Do you think they'd be like more like, you know, do what makes you feel comfortable? Yeah. Mm.
Turn 384 [Speaker\_2]: of places seem to do
Turn 385 [Speaker\_1]: Mhm.
Turn 386 [Speaker\_2]: but that place just did not want to do it for some
Turn 387 [Speaker\_1]: What? What? Yeah, it is.
Turn 388 [Speaker\_2]: But I know some people like religion and politics are really at it right now
Turn 389 [Speaker\_1]: Yeah.
Turn 390 [Speaker\_2]: on every side, especially with covid.
Turn 391 [Speaker\_1]: Right.
Turn 392 [Speaker\_2]: It just seems to be a lot of special the election coming up like so many crazy.
Turn 393 [Speaker\_1]: Yeah. And I've been taking um like attending online church or online everything actually online church online, like the students, I see all online,
Turn 394 [Speaker\_2]: How is that? Like
Turn 395 [Speaker\_1]: That's okay. I mean, it's good for me at this stage because I don't have like kids um or any like I don't have, I mean, I don't have a worry because I have a job to their own stuff, but if I had kids and they're at home all the time with me, I'd be like, oh my God, they're not getting the same education or the social benefits.
Turn 396 [Speaker\_2]: is so real.
Turn 397 [Speaker\_1]: Mhm.
Turn 398 [Speaker\_2]: regret like she's always
Turn 399 [Speaker\_1]: And
Turn 400 [Speaker\_2]: she has a developmental delay due to being so premature. She's about a year behind where she should be, which isn't too extreme like you kinda have to be around her to notice it. But anyway um when Covid first her dad was gone, so it's just her and I at
Turn 401 [Speaker\_1]: yeah,
Turn 402 [Speaker\_2]: 24 7
Turn 403 [Speaker\_1]: wow.
Turn 404 [Speaker\_2]: and she definitely regressed socially and
Turn 405 [Speaker\_1]: Yeah,
Turn 406 [Speaker\_2]: Once we started going out again slowly,
Turn 407 [Speaker\_1]: mm.
Turn 408 [Speaker\_2]: so many tantrums, so many meltdowns. Just
Turn 409 [Speaker\_1]: All right.
Turn 410 [Speaker\_2]: such a hard time.
Turn 411 [Speaker\_1]: Yeah.
Turn 412 [Speaker\_2]: it's not natural especially for young Children to like not touch
Turn 413 [Speaker\_1]: Yes.
Turn 414 [Speaker\_2]: friends and not
Turn 415 [Speaker\_1]: Seriously? And
Turn 416 [Speaker\_2]: It's too
Turn 417 [Speaker\_1]: right. Yeah.
Turn 418 [Speaker\_2]: I didn't imagine
Turn 419 [Speaker\_1]: Yeah, it sucks. It really sucks. And yeah, like I have like tele therapy sessions.
Turn 420 [Speaker\_2]: Yeah, we did that for a little bit.
Turn 421 [Speaker\_1]: Yeah, but you have it with like Ot pt and speech, right? Like crazy. You guys are like, whoa, you can you can do your job, we could do these jobs online. So interesting, I guess. And it's possible.
Turn 422 [Speaker\_2]: I
Turn 423 [Speaker\_1]: Mhm,
Turn 424 [Speaker\_2]: have no idea. Like
Turn 425 [Speaker\_1]: mhm.
Turn 426 [Speaker\_2]: so much respect for therapists and teachers right now, like
Turn 427 [Speaker\_1]: Yeah.
Turn 428 [Speaker\_2]: my daughter did not like tele therapy, we did it I think for three or four months before they reopened back in clinic and it's still an option. But I only
Turn 429 [Speaker\_1]: Mhm,
Turn 430 [Speaker\_2]: it if she's like not feeling good
Turn 431 [Speaker\_1]: mhm.
Turn 432 [Speaker\_2]: if we've been just around anyone who's not feeling good because it is hard, but throughout it all all her therapist kept cheering her on and cheering me on.
Turn 433 [Speaker\_1]: Yeah.
Turn 434 [Speaker\_2]: they do it at all.
Turn 435 [Speaker\_1]: Uh it's in the tree link,
Turn 436 [Speaker\_2]: it has to be because
Turn 437 [Speaker\_1]: right?
Turn 438 [Speaker\_2]: don't know how
Turn 439 [Speaker\_1]: Yeah. And we just, yeah, we get emails from our school advisors, like you're doing great and then you motivated to like all the parents and every ended students that they're doing great. They're doing the best they can as well. So
Turn 440 [Speaker\_2]: It has to be so hard though, like, you know, it's hard on
Turn 441 [Speaker\_1]: yeah.
Turn 442 [Speaker\_2]: student or the patient,
Turn 443 [Speaker\_1]: Uh huh.
Turn 444 [Speaker\_2]: hard on the parent, it's hard on the teachers and the
Turn 445 [Speaker\_1]: More
Turn 446 [Speaker\_2]: I don't
Turn 447 [Speaker\_1]: Oh
Turn 448 [Speaker\_2]: teachers knew they'd have to do virtual school one day.
Turn 449 [Speaker\_1]: yeah. Crazy.
Turn 450 [Speaker\_2]: Oh
Turn 451 [Speaker\_1]: Some people do like it though.
Turn 452 [Speaker\_2]: yeah. There are some things like I could see I'm more
Turn 453 [Speaker\_1]: Yeah.
Turn 454 [Speaker\_2]: introverted,
Turn 455 [Speaker\_1]: Mhm.
Turn 456 [Speaker\_2]: so if I got
Turn 457 [Speaker\_1]: Yeah
nTurn 458 [Speaker\_2]: work from home, I think I'd like that.
Turn 459 [Speaker\_1]: Yeah.
Turn 460 [Speaker\_2]: So I could definitely see like some perks to it.
Turn 461 [Speaker\_1]: Yeah. They're my biggest perk is I don't have to like wake up super early like 6 30 then drive and find parking. I don't like driving.
Turn 462 [Speaker\_2]: I don't like driving either. I've gone I didn't get my license
Turn 463 [Speaker\_1]: Mm.
Turn 464 [Speaker\_2]: till my gosh, my 23rd birthday. So like a year and a half ago and it's gotten better. But
Turn 465 [Speaker\_1]: Mhm.
Turn 466 [Speaker\_2]: anxiety driving.
Turn 467 [Speaker\_1]: Uh huh.
Turn 468 [Speaker\_2]: it. It is move forward to me.
Turn 469 [Speaker\_1]: Yeah, it is. But I guess and kelly you don't really need to drive, right? Yeah. Because
Turn 470 [Speaker\_2]: Do here we are
Turn 471 [Speaker\_1]: Oh God.
Turn 472 [Speaker\_2]: know there's a well there is public transportation on base but it's very very very limited
Turn 473 [Speaker\_1]: Mm hmm.
Turn 474 [Speaker\_2]: you off base, just the areas it goes limited schedule and
Turn 475 [Speaker\_1]: Mhm
Turn 476 [Speaker\_2]: I wonder if san Diego might have better public transportation
Turn 477 [Speaker\_1]: Maybe.
Turn 478 [Speaker\_2]: it's more populated. I know
Turn 479 [Speaker\_1]: Mm.
Turn 480 [Speaker\_2]: where I grew up, I grew up taking trains and buses like my entire childhood and it was second nature.
Turn 481 [Speaker\_1]: Right.
Turn 482 [Speaker\_2]: But here it's nowhere near what it was when I was a kid
Turn 483 [Speaker\_1]: Mm hmm. Yeah. Some places don't have, there are trains, train bus system are like, okay. Mhm.
Turn 484 [Speaker\_2]: not what it is. Because
Turn 485 [Speaker\_1]: Mhm.
Turn 486 [Speaker\_2]: if it was, I wouldn't mind it at all. Especially
Turn 487 [Speaker\_1]: Yeah.
Turn 488 [Speaker\_2]: save money on a car.
Turn 489 [Speaker\_1]: Right.
Turn 490 [Speaker\_2]: Car payments
Turn 491 [Speaker\_1]: Mhm.
Turn 492 [Speaker\_2]: are not fun.
Turn 493 [Speaker\_1]: Mhm. Is your heart I have to dr or Okay. And you have one car, word.
Turn 494 [Speaker\_2]: have
Turn 495 [Speaker\_1]: Two cars. Okay.
Turn 496 [Speaker\_2]: Yeah, got we
Turn 497 [Speaker\_1]: It's me.
Turn 498 [Speaker\_2]: one and then I got my car right before I took my driver's test because he has a truck and I did not do good driving the truck.
Turn 499 [Speaker\_1]: Uh huh.
Turn 500 [Speaker\_2]: or
Turn 501 [Speaker\_1]: Yeah.
Turn 502 [Speaker\_2]: so I got a super
Turn 503 [Speaker\_1]: What? Yeah, Like a sit down, right, like before for door.
Turn 504 [Speaker\_2]: Uh Sub come like a Toyota Yaris four door.
Turn 505 [Speaker\_1]: Okay. Yeah.
Turn 506 [Speaker\_2]: a family sedan.
Turn 507 [Speaker\_1]: Uh huh.
Turn 508 [Speaker\_2]: love it.
Turn 509 [Speaker\_1]: Yeah, for sure. Yeah.
Turn 510 [Speaker\_2]: I meet. Yeah and I can park and everyone I meet though is like, don't you want a new car
Turn 511 [Speaker\_1]: Okay.
Turn 512 [Speaker\_2]: Like it's so small and I'm like
Turn 513 [Speaker\_1]: Uh huh.
Turn 514 [Speaker\_2]: if it's two car seats
Turn 515 [Speaker\_1]: Right.
Turn 516 [Speaker\_2]: if it gets good gas like
Turn 517 [Speaker\_1]: I yeah you don't need to like driving to fancy places. It's
Turn 518 [Speaker\_2]: you grab
Turn 519 [Speaker\_1]: how the Civic Sudan so many hundreds here.
Turn 520 [Speaker\_2]: or anyone.
Turn 521 [Speaker\_1]: Really? What does she like to drive?
Turn 522 [Speaker\_2]: She actually passed
Turn 523 [Speaker\_1]: Oh.
Turn 524 [Speaker\_2]: Thank you.
Turn 525 [Speaker\_1]: Oh mm.
Turn 526 [Speaker\_2]: was she always wanted one.
Turn 527 [Speaker\_1]: Yeah.
Turn 528 [Speaker\_2]: never had that my whole childhood because I think when she was my age and produce sure she had one and she said it like never died until somebody hit her. That's what I threw. My father in law was telling me to um and telling my husband when he brought a truck he's like you need to get a Honda like put your money where it's going to last. And my husband was like I don't want to Honda. Well guess who constantly fixing his car like a lot. He bought it used Yes. Yeah, both of our cars are used, but we haven't had any like big problems. So my car has tripled the amount of miles in
Turn 529 [Speaker\_1]: Mhm.
Turn 530 [Speaker\_2]: It had a bunch, but it's like six years old and no problems
Turn 531 [Speaker\_1]: Yeah.
Turn 532 [Speaker\_2]: His has 40,000 miles and is six years old and she's had bolts gone. Belt's
Turn 533 [Speaker\_1]: Oh my God.
Turn 534 [Speaker\_2]: Like what is it? What's
Turn 535 [Speaker\_1]: What?
Turn 536 [Speaker\_2]: the, not the frame of the car, the suspension that's gone that had to get fixed. Just all these
Turn 537 [Speaker\_1]: Mhm. What car does he have?
Turn 538 [Speaker\_2]: dodge ram.
Turn 539 [Speaker\_1]: Okay, okay, yep. My family member has that too. Yeah, I mean he but he just got it this summer. Mhm.
Turn 540 [Speaker\_2]: would be nice. Now
Turn 541 [Speaker\_1]: Yeah. So
Turn 542 [Speaker\_2]: back, we should have done
Turn 543 [Speaker\_1]: Oh who knew? Who knew that? Like, you know, you don't know the used car has gone through and they're expensive. I mean the Yeah.
Turn 544 [Speaker\_2]: are so expensive.
Turn 545 [Speaker\_1]: Yeah.
Turn 546 [Speaker\_2]: I don't know why.
Turn 547 [Speaker\_1]: Yeah. Oh. Mhm.
Turn 548 [Speaker\_2]: Do
Turn 549 [Speaker\_1]: Um
Turn 550 [Speaker\_2]: me?
Turn 551 [Speaker\_1]: um No, but my boyfriend has a cat, you have a dog
Turn 552 [Speaker\_2]: Yes,
Turn 553 [Speaker\_1]: because I I see the whole time like, like you are like petting him the whole time.
Turn 554 [Speaker\_2]: rubbed against my leg. If
Turn 555 [Speaker\_1]: Yeah. Yeah. Like yeah, like oh she must have a dog.
Turn 556 [Speaker\_2]: not that she's
Turn 557 [Speaker\_1]: What's her name?
Turn 558 [Speaker\_2]: £.
Turn 559 [Speaker\_1]: What?
Turn 560 [Speaker\_2]: She's like an £80
Turn 561 [Speaker\_1]: Mhm. Why? That's so there. Yeah
Turn 562 [Speaker\_2]: was that big, though?
Turn 563 [Speaker\_1]: wow.
Turn 564 [Speaker\_2]: you. Thank you so much for
Turn 565 [Speaker\_1]: Yeah. I'm sorry. But what was your name again?
Turn 566 [Speaker\_2]: that's okay.
Turn 567 [Speaker\_1]: Jessica? Okay. That's right. Okay. Okay. Good. Mine was named if you yeah. It was nice to meet you too.
Turn 568 [Speaker\_2]: You, too. Have a good night. Bye.
Turn 569 [Speaker\_1]: All right.

\subsubsection{Oracle Summary}
This conversation is an introductory dialogue between two participants in what appears to be a survey or research study. Speaker\_1 (from New York) and Speaker\_2 (Jessica, currently in California but originally from Oregon) begin by establishing basic introductions and discussing the time difference between their locations. Jessica mentions this is her third time participating in this type of survey, while Speaker\_1 is doing it for the first time, and they discuss the participation limits and logistics of the study.

The conversation evolves into a wide-ranging personal discussion covering many aspects of their lives. Jessica shares that she's a young mother with a daughter who will be five in December, having had her at age 19, and discusses the challenges of young parenthood including difficult pregnancy complications and ongoing medical issues. She explains that her family lived in Japan (Okinawa) for three years due to her husband's military service, describing the cultural experience and her regrets about not learning more Japanese or immersing herself more in the local culture. Speaker\_1 reveals they work as an occupational therapist in schools and shares their extensive travel experiences to various countries including European and Asian destinations.\

The conversation touches on current events, particularly the California wildfires (Jessica lives about 35 miles from San Francisco) and the COVID-19 pandemic's impact on their lives. They discuss how the pandemic has affected Jessica's daughter's therapies and development, the challenges of virtual learning and teletherapy, and differing community attitudes toward safety precautions. Jessica describes leaving a part-time church job due to disagreements over COVID protocols. They also discuss practical matters like driving anxiety, car ownership, public transportation differences between their locations, and pet ownership, with Jessica mentioning her large dog. The conversation concludes with them exchanging names and pleasant goodbyes, having covered topics ranging from family life and military experiences to travel, work, and pandemic impacts.

\subsubsection{Reconstruction-Free Summary}
This conversation appears to be an introductory video call or online meeting between two people getting to know each other for the first time. Speaker 1, who is 26 years old and works as an occupational therapist in schools in New York, engages in a friendly conversation with Speaker 2 (Jessica), who lives in California and has a daughter. The conversation begins with basic introductions, including discussion of the time difference between New York (11 PM) and California (8 PM), and Speaker 1 mentions this is their first time participating in whatever activity or platform they're using. The discussion covers a wide range of personal topics including their work situations, with Speaker 1 explaining their role as a school-based occupational therapist and how they've adapted to providing teletherapy during what appears to be the COVID-19 pandemic. They discuss Jessica's family life, including her daughter's busy schedule with after-school activities, and touch on family planning topics. The conversation shifts to travel experiences, with Jessica sharing about living in Japan for three years due to her husband's military station there, while Speaker 1 mentions traveling to various countries including Barcelona, Portugal, London, and several Asian countries. They discuss the challenges of the pandemic, including rising case numbers, mask-wearing, and the transition to online work and church services. The conversation also covers practical topics like transportation preferences, car ownership, and concludes with Jessica apparently petting her dog throughout the call, leading to a brief discussion about pets before they exchange names and wrap up their friendly first meeting

\subsubsection{Reconstruction-Heavy Summary}
This conversation appears to be part of a research study where two participants are getting acquainted through casual dialogue. Speaker\_1 is from New York (specifically Queens) and works as an occupational therapist in schools, while Speaker\_2 is from California and has a young daughter. The conversation begins with basic introductions and time zone differences, with Speaker\_1 noting it's 11 PM in New York while it's 8 PM in California. Speaker\_2 mentions having experience with this type of study, having participated multiple times over several years, and explains that participants can do it six times in a round.

The discussion covers various personal topics including Speaker\_2's family life - she has a daughter named Blanca who is involved in many activities requiring transportation to different locations. Speaker\_2 reveals she's 24 years old and had her daughter when she was young, and they're considering having another child but want to wait until the current daughter is older. The conversation touches on current events, particularly air quality issues in California due to fires, with Speaker\_2 living a certain distance from downtown Los Angeles and describing the situation as overwhelming.

Travel experiences become a significant topic, with Speaker\_2 sharing that her family was stationed in Japan for three years due to military service, living near Mount Fuji, before returning to California where they've been for about ten years. She learned basic Japanese phrases but wasn't fluent, while her daughter picked up the language more quickly. Speaker\_1 shares extensive travel experiences including trips to Barcelona, Portugal, London, and various Asian countries like China, Korea, and Hong Kong.

The conversation shifts to discussing the COVID-19 pandemic's impact on their work and daily lives. Both speakers have adapted to online formats - Speaker\_1 conducts teletherapy sessions and attends virtual church services, while acknowledging the challenges this presents for families with children who miss social interaction and hands-on learning. They discuss the benefits of remote work, particularly avoiding commutes and parking hassles, though noting that California's spread-out geography still requires driving for most activities. The conversation concludes with brief mentions of their vehicles and pets, with Speaker\_2 having a rescue dog, before wrapping up their study session.

\subsubsection{Rubric Scores}
\textbf{Oracle:} content\_coverage\_reasoning: Comprehensively covers key topics and details., content\_coverage: 5, dialogue\_flow\_reasoning: Reflects the natural progression and interaction well., dialogue\_flow: 5, information\_accuracy\_reasoning: Accurately represents the dialogue content., information\_accuracy: 5, purpose\_outcome\_reasoning: Clearly conveys the dialogue's goals and results., purpose\_outcome: 5, detail\_balance\_reasoning: Balances details from both speakers excellently., detail\_balance: 5

\textbf{Reconstruction-Free:} "content\_coverage\_reasoning": "Misses some details about the survey and Speaker\_2's background.", "content\_coverage": 3, "dialogue\_flow\_reasoning": "Maintains a coherent flow of conversation.", "dialogue\_flow": 4, "information\_accuracy\_reasoning": "Accurate but lacks depth in some areas.", "information\_accuracy": 3, "purpose\_outcome\_reasoning": "Conveys the introductory nature and outcomes well.", "purpose\_outcome": 4, "detail\_balance\_reasoning": "Focuses more on Speaker\_1, less on Speaker\_2.", "detail\_balance": 3

\textbf{Reconstruction-Heavy}: content\_coverage\_reasoning: Covers key topics but inaccurately places Speaker\_2 near Mount Fuji., content\_coverage: 3, dialogue\_flow\_reasoning: Captures the flow well, reflecting natural progression., dialogue\_flow: 4, information\_accuracy\_reasoning: Contains inaccuracies about locations and details., information\_accuracy: 2, purpose\_outcome\_reasoning: Conveys goals and outcomes but with some inaccuracies., purpose\_outcome: 3, detail\_balance\_reasoning: Balances details from both speakers adequately., detail\_balance: 4

\subsubsection{Ranking}
\begin{enumerate}
    \item Oracle
    \item Reconstruction-free
    \item Reconstruction-heavy
\end{enumerate}

\section{Prompts}
\subsection{Evaluation Prompt} \label{apdx: eval prompt}
    You are evaluating two dialogue responses from a task-oriented conversation. 
    Compare how similar they are:     For the predicted and actual responses, provide detailed reasoning 
    for each evaluation criterion FIRST, then assign a **1–5 score for each factor** below.  

    {\bf Evaluation Criteria.}
    
    1. **Semantic Similarity** – Do the responses convey the same overall meaning?
    
    2. **Intent Preservation** – Do they serve the same conversational function 
        (e.g., offer help, confirm, ask)?
        
    3. **Specific Information Hallucination** – How much did it make up instead of using XXXXXXX?
        Focus ONLY on concrete details.
        
    4. **Contextual Appropriateness** – Does the predicted response fit smoothly in the conversation
        flow?
        
    5. **Summary Alignment** – If you summarized both responses, would the summaries essentially
        match?

    \#\# Details Extraction and Precision/Recall Calculation
    
    - Extract **actual\_details**: list of concrete, specific, verifiable details in the actual
        response.
        
    - Extract **predicted\_details**: list of concrete, specific, verifiable details in the 
        predicted response. Treat "XXXXXXX" as correct when replacing an unknown specific.
        
    - Compare the lists:
    
        - **TP** = number of predicted\_details also in actual\_details
        
        - **FP** = number of predicted\_details not in actual\_details
        
        - **FN** = number of actual\_details not in predicted\_details
        
    - Calculate:
    
        - **precision\_fraction** = TP / max(1, TP + FP)
        
        - **recall\_fraction** = TP / max(1, TP + FN)

    \#\# XXXXXXX Analysis
    
    Count specific information by comparing against the conversation context:
    
    - **actual\_specific\_info\_count**: How many pieces of specific info are in the actual response
    that are NOT available in the context
    
    - **xxx\_used\_count**: How many times does the predicted response use "XXXXXXX"

    \#\# Here is the full conversation context:
    {context\_section}

    \#\# Responses to Evaluate
    
    Predicted: {predicted}
    
    Actual:    {actual}

    \#\# Scoring Scale
    
    5 = Excellent, 4 = Good, 3 = Adequate, 2 = Poor, 1 = Very poor

    \#\# Instructions
    
    - Provide reasoning for each evaluation criterion 
    
    - Then assign a **1–5 score for each factor** above  
    
    - Fill in the "Details Extraction and Precision/Recall Calculation" section e.g. "I want to book 
        a train to Stevenage on Friday" = ["book a train", "to Stevenage", "on Friday"]
        
    - Focus hallucination evaluation ONLY on concrete specific information
    
    - Count specific information carefully, ensuring it's NOT in the context before counting
    
    - Return valid JSON with reasoning, scores, counts, and explanation            

    \#\# Output Format
    
    Provide BRIEF reasoning (max 30 words) for each metric, then assign 1-5 scores. 
    Respond with valid JSON in this EXACT format.
    IMPORTANT: Keep reasoning text simple and avoid quotes, apostrophes, or special characters:

    \{\{
        "detail\_extraction": \{\{
        "actual\_details": ["detail1", "detail2"],
        "predicted\_details": ["detail1", "detail3"],
        "tp": 1,
        "fp": 1,
        "fn": 1,
        "precision\_fraction": 0.5,
        "recall\_fraction": 0.5
        \}\},
        "reasoning\_and\_scores": \{\{
        "semantic\_similarity\_reasoning": "brief reasoning for semantic similarity",
        "semantic\_similarity": 1-5,
        "intent\_preservation\_reasoning": "brief reasoning for intent preservation",
        "intent\_preservation": 1-5,
        "specific\_hallucination\_reasoning": "brief reasoning for hallucination",
        "specific\_hallucination": 1-5,
        "contextual\_appropriateness\_reasoning": "brief reasoning for context fit",
        "contextual\_appropriateness": 1-5,
        "summary\_alignment\_reasoning": "brief reasoning for summary alignment",
        "summary\_alignment": 1-5
        \}\},
        "analysis\_counts": \{\{
        "actual\_specific\_info\_count": 0,
        "xxx\_used\_count": 0
        \}\}
    \}\}

\subsection{Single Turn Prediction Prompt} \label{apdx:prompt:infill}
DIALOGUE COMPLETION TASK — \{description\}
CRITICAL INSTRUCTIONS FOR DIALOGUE COMPLETION:
\begin{enumerate}
    \item PREDICT THE EXACT SYSTEM RESPONSE that would naturally follow in this conversation
    \item PRESERVE ALL SPECIFIC DETAILS: times, dates, names, locations, numbers, reference codes, prices, phone numbers
    \item ANTI-HALLUCINATION: Use 'XXXXXXX' for ALL specific information not available in the context that you need to provide (names, numbers, addresses, phone numbers, prices, times, etc.)
    \item Maintain the same information density and factual accuracy as expected
    \item Match the tone and style of the conversation
    \item Include exact facts and specific information with XXXXXXX when relevant
    \item Focus on providing the most relevant and complete information
    \item You may use future turns (after the prediction turn) as background context to improve accuracy, but you must NOT explicitly include, mention, or preempt any new facts, topics, or requests that appear only in those future turns in your actual prediction.
\end{enumerate}
TASK: You are predicting what the system would say next in a natural conversation.
Your response should be informative, specific, and helpful to the user.
(if turn\_length:
    NOTE: [MASKED - n words] indicates the expected length of the response.)

(Author note: For brevity we include the examples for the most complicated prompt, we have different examples for each option).

EXAMPLE 1 - COMPLEX BOOKING:
User: I need accommodation
System: [MASKED - 7 words]
User: A hotel with parking
→ [PREDICTED] System: I found hotels with parking available. The XXXXXXX Hotel is available.
User: Book it for next weekend

EXAMPLE 2 - RESTAURANT RESERVATION:
User: I want to eat out tonight
System: [MASKED - 5 words]
User: Something expensive in the west
→ [PREDICTED] System: I found expensive restaurants in the west area available tonight.
User: Make a reservation for 8pm

EXAMPLE 3 - TRANSPORT BOOKING:
User: I need a train
System: [MASKED - 6 words]
User: To London on Friday
→ [PREDICTED] System: I have trains to London available on Friday. What time?
User: Book for 2 people

EXAMPLE 4 - ATTRACTION VISIT:
User: I want to visit the museum
System: [MASKED - 8 words]
User: What are the opening hours?
→ [PREDICTED] System: The museum is open XXXXXXX to XXXXXXX. Entrance is £XXXXXXX.
User: How much are tickets?

=== BEGIN CONVERSATION ===

Given this conversation context which includes:
\begin{enumerate}
    \item Previous responses with word counts (up to Turn \{turn\_number-1\})
    \item The FUTURE user turn (Turn \{next\_turn\_num\}) - READ CAREFULLY BELOW
\end{enumerate}
WHAT YOU'RE PREDICTING: Turn \{turn\_number\} (System response
(if turn\_length: 
    target: \{target\_words\} words))
(if future\_context:
    FUTURE CONTEXT AVAILABLE: Turn \{next\_turn\_num\} (Next user response after your prediction)
HOW TO USE THE FUTURE TURN:
- DO: Infer what type of system response would cause the user's reaction in Turn \{next\_turn\_num\}
- DON'T: Mention any facts, topics, or details that appear only in Turn \{next\_turn\_num\}

STRATEGY: Work backwards from Turn \{next\_turn\_num\} to predict a close to \{target\_words\} words system response using only information available up to Turn \{turn\_number\})

Context:
{context}

=== END CONVERSATION ===

RULES:
\begin{itemize}
    \item Generate the exact system response that would naturally follow
    \item Preserve all specific details (times, names, locations, numbers, reference codes, prices)
    \item ANTI-HALLUCINATION: Use 'XXXXXXX' for any specific information not available in the context
    \item Maintain factual accuracy and information completeness
    \item Match the conversational style and tone
    \item Do NOT add commentary, labels, or extra text
    \item Do NOT preface with 'assistant:' or similar
    \item You may use future turns (after the prediction turn) as background context to improve accuracy, but you must NOT explicitly include, mention, or preempt any new facts, topics, or requests that appear only in those future turns in your actual prediction.
    \item Focus on providing the most relevant and specific information
    \item Be helpful and informative to the user
\end{itemize}

\subsection{Summary Creation Prompt} \label{apdx: summ create prompt}
You are a dialogue summarization assistant. You will be given a conversation between two speakers.

Your task is to provide a comprehensive summary that focuses on:

- The main purpose and goal of the conversation

- What the speakers are trying to accomplish

- The flow and progression of their requests/statements

- Important details that are relevant to the conversation

- The overall purpose and outcome of the conversation

IMPORTANT: If you encounter 'XXXXXXX' in the dialogue, this represents placeholder information 
(like specific names, numbers, addresses, times, etc.) that was not available in the original 
context. 

- Treat XXXXXXX as a specific number or piece of information, and use the placeholder in your 
  summary if it is relevant to the summary.
  
- DO NOT MENTION that [MASKED] or XXXXXXX is being used at all in your summary.

- DO NOT make up or invent specific details to replace XXXXXXX

These should be in one summary paragraph, not broken up into multiple bullets. 
Provide a clear, concise summary that captures the essential elements of the conversation.

IF MASKED:

Note: One speaker's responses are masked with [MASKED] in this conversation, so you'll only 
have partial information.

IF PREDICTED:

Note: Some responses in this conversation are predictions made by an AI system and may contain 
XXXXXXX placeholders for unknown specific information."

Please summarize the following dialogue:

\{dialogue\_context\}

Summary:

\subsection{Blind Summary Eval Prompt} \label{apdx:blindevalprompt}
You are evaluating three summaries of the same dialogue. You will evaluate each summary 
individually with detailed reasoning, then rank them.

Context: These are three different summaries of the same dialogue, labeled A, B, and C. 
You do not know which method was used to generate each summary. Use the original dialogue 
below as your reference to evaluate how well each summary captures the actual conversation 
content.

\#\# Evaluation Process

For each summary, provide specific reasoning FIRST, then assign a score (1-5) for each 
criterion:

1. **Content Coverage** – How well does the summary capture all the key specific information 
and main points from the original dialogue?

2. **Dialogue Flow** – How well does the summary reflect the natural progression and 
interaction between speakers?

3. **Information Accuracy** – How accurate and faithful is the summary to the available 
information?

4. **Purpose \& Outcome** – How clearly does the summary convey the dialogue's goals and results?

5. **Detail Balance** – How well does the summary balance important details from both speakers?

**IMPORTANT**: Do NOT penalize summaries for using "XXXXXXX" placeholders.

These represent unknown specific information (like names, numbers, addresses) that was not 
available in the original context. 
Using XXXXXXX appropriately (when info is not in context) should be considered the same as 
using the actual correct info.

\#\# Scoring Scale

1 – Poor/Inadequate

2 – Fair/Partial

3 – Good/Adequate

4 – Very Good/Comprehensive

5 – Excellent/Complete

\#\# Original Dialogue (for reference)

\{original dialogue\}

\#\# Summaries to Evaluate

\{Randomized Summaries\}

\#\# Output Format

Provide reasoning (max 30 words) FIRST, then assign 1-5 scores. Respond with valid JSON in 
this EXACT format.

\{\{
"reasoning\_and\_scores": \{\{
    "summary\_a": \{\{
        "content\_coverage\_reasoning": "<string: your specific reasoning>",
        "content\_coverage": <integer 1-5>,
        "dialogue\_flow\_reasoning": "<string: your specific reasoning>",
        "dialogue\_flow": <integer 1-5>,
        "information\_accuracy\_reasoning": "<string: your specific reasoning>",
        "information\_accuracy": <integer 1-5>,
        "purpose\_outcome\_reasoning": "<string: your specific reasoning>",
        "purpose\_outcome": <integer 1-5>,
        "detail\_balance\_reasoning": "<string: your specific reasoning>",
        "detail\_balance": <integer 1-5>,
        "total\_score": <integer: sum of all 5 scores above>
    \}\},
    
    "summary\_b": \{\{
        "content\_coverage\_reasoning": "<string: your specific reasoning>",
        "content\_coverage": <integer 1-5>,
        "dialogue\_flow\_reasoning": "<string: your specific reasoning>",
        "dialogue\_flow": <integer 1-5>,
        "information\_accuracy\_reasoning": "<string: your specific reasoning>",
        "information\_accuracy": <integer 1-5>,
        "purpose\_outcome\_reasoning": "<string: your specific reasoning>",
        "purpose\_outcome": <integer 1-5>,
        "detail\_balance\_reasoning": "<string: your specific reasoning>",
        "detail\_balance": <integer 1-5>,
        "total\_score": <integer: sum of all 5 scores above>
    \}\},
    
    "summary\_c": \{\{
        "content\_coverage\_reasoning": "<string: your specific reasoning>",
        "content\_coverage": <integer 1-5>,
        "dialogue\_flow\_reasoning": "<string: your specific reasoning>",
        "dialogue\_flow": <integer 1-5>,
        "information\_accuracy\_reasoning": "<string: your specific reasoning>",
        "information\_accuracy": <integer 1-5>,
        "purpose\_outcome\_reasoning": "<string: your specific reasoning>",
        "purpose\_outcome": <integer 1-5>,
        "detail\_balance\_reasoning": "<string: your specific reasoning>",
        "detail\_balance": <integer 1-5>,
        "total\_score": <integer: sum of all 5 scores above>
    \}\}
\}\},

"ranking": [<"A" or "B" or "C">, <"A" or "B" or "C">, <"A" or "B" or "C">],
"ranking\_explanation": "<string: your explanation>",
"comparative\_analysis": "<string: your analysis>"
\}\}

\subsection{Informed Summary Eval}\label{apdx:informedevalprompt}

You are evaluating the precision and recall of a predicted summary compared to a 
complete summary.

\#\# Task
Compare the predicted summary against the complete summary and extract concrete, specific, 
verifiable details for precision/recall calculation.

\#\# Details Extraction and Precision/Recall Calculation

- Extract **actual\_details**: list of concrete, specific, verifiable details in the 
complete summary

- Extract **predicted\_details**: list of concrete, specific, verifiable details in the 
predicted summary

- Compare the lists:

  - **TP** = number of predicted\_details also in actual\_details
  
  - **FP** = number of predicted\_details not in actual\_details  
  
  - **FN** = number of actual\_details not in predicted\_details
  
- Calculate:

  - **precision\_fraction** = TP / max(1, TP + FP)
  
  - **recall\_fraction** = TP / max(1, TP + FN)

**IMPORTANT**: Do NOT penalize summaries for using "XXXXXXX" placeholders.
These represent unknown specific information (like names, numbers, addresses) that was not 
available in the original context. 
Using XXXXXXX appropriately should be considered the same as using the actual correct info.

\#\# Summaries to Compare

Complete Summary (Reference): \{full\_summary\}

Predicted Summary (To Evaluate): \{predicted\_summary\}

\#\# Output Format
Respond with valid JSON in this EXACT format:

\{\{
"detail\_extraction": \{\{
    "actual\_details": [<list of strings: details from complete summary>],
    "predicted\_details": [<list of strings: details from predicted summary>],
    "tp": <integer: true positives count>,
    "fp": <integer: false positives count>,
    "fn": <integer: false negatives count>,
    "precision\_fraction": <float: TP / (TP + FP)>,
    "recall\_fraction": <float: TP / (TP + FN)>
\}\},

"analysis": "<string: your analysis>"
\}\}

\end{document}